\documentclass[acmsmall]{acmart}

\usepackage{multirow}
\usepackage{color, soul}
\usepackage{makecell}
\usepackage{hyperref}
\usepackage{CJKutf8}

\sethlcolor{yellow}
\soulregister\cite7
\soulregister\ref7

\AtBeginDocument{%
  \providecommand\BibTeX{{%
    \normalfont B\kern-0.5em{\scshape i\kern-0.25em b}\kern-0.8em\TeX}}}

\setcopyright{acmcopyright}
\copyrightyear{2018}
\acmYear{2018}
\acmDOI{10.1145/1122445.1122456}

\acmJournal{JACM}
\acmVolume{37}
\acmNumber{4}
\acmArticle{111}
\acmMonth{8}

\begin{document}

\title{Deep Person Generation: A Survey from the Perspective of Face, Pose and Cloth Synthesis}

\author{Tong Sha}
\email{tongsha@buaa.edu.cn}
\affiliation{%
  \institution{Beihang University}
  \city{Beijing}
  \country{China}
}

\author{Wei Zhang}
\email{wzhang.cu@gmail.com}
\affiliation{%
  \institution{JD AI Research}
  \city{Beijing}
  \country{China}
}

\author{Tong Shen}
\email{tshen.st@outlook.com}
\affiliation{%
  \institution{JD AI Research}
  \city{Beijing}
  \country{China}
}

\author{Zhoujun Li}
\email{lizj@buaa.edu.cn}
\affiliation{%
  \institution{Beihang University}
  \city{Beijing}
  \country{China}
}

\author{Tao Mei}
\email{tmei@live.com}
\affiliation{%
  \institution{JD AI Research}
  \city{Beijing}
  \country{China}
}

\renewcommand{\shortauthors}{Sha and Zhang, et al.}

\begin{abstract}
Deep person generation has attracted extensive research attention due to its wide applications in virtual agents, video conferencing, online shopping and art/movie production. With the advancement of deep learning, visual appearances (face, pose, cloth) of a person image can be easily generated on demand. 
In this survey, we first summarize the scope of person generation, and then systematically review recent progress and technical trends in identity-preserving deep person generation, covering three major tasks: \textit{talking-head generation} (face), \textit{pose-guided person generation} (pose) and \textit{garment-oriented person generation} (cloth). 
More than two hundred papers are covered for a thorough overview, and the milestone works are highlighted to witness the major technical breakthrough. 
Based on these fundamental tasks, many applications are investigated, e.g., virtual fitting, digital human, generative data augmentation.
We hope this survey could shed some light on the future prospects of identity-preserving deep person generation, and provide a helpful foundation for full applications towards the digital human.
\end{abstract}



\maketitle
\section{Introduction}

\begin{figure}[t]
	\centering
	\includegraphics[width=0.9\textwidth]{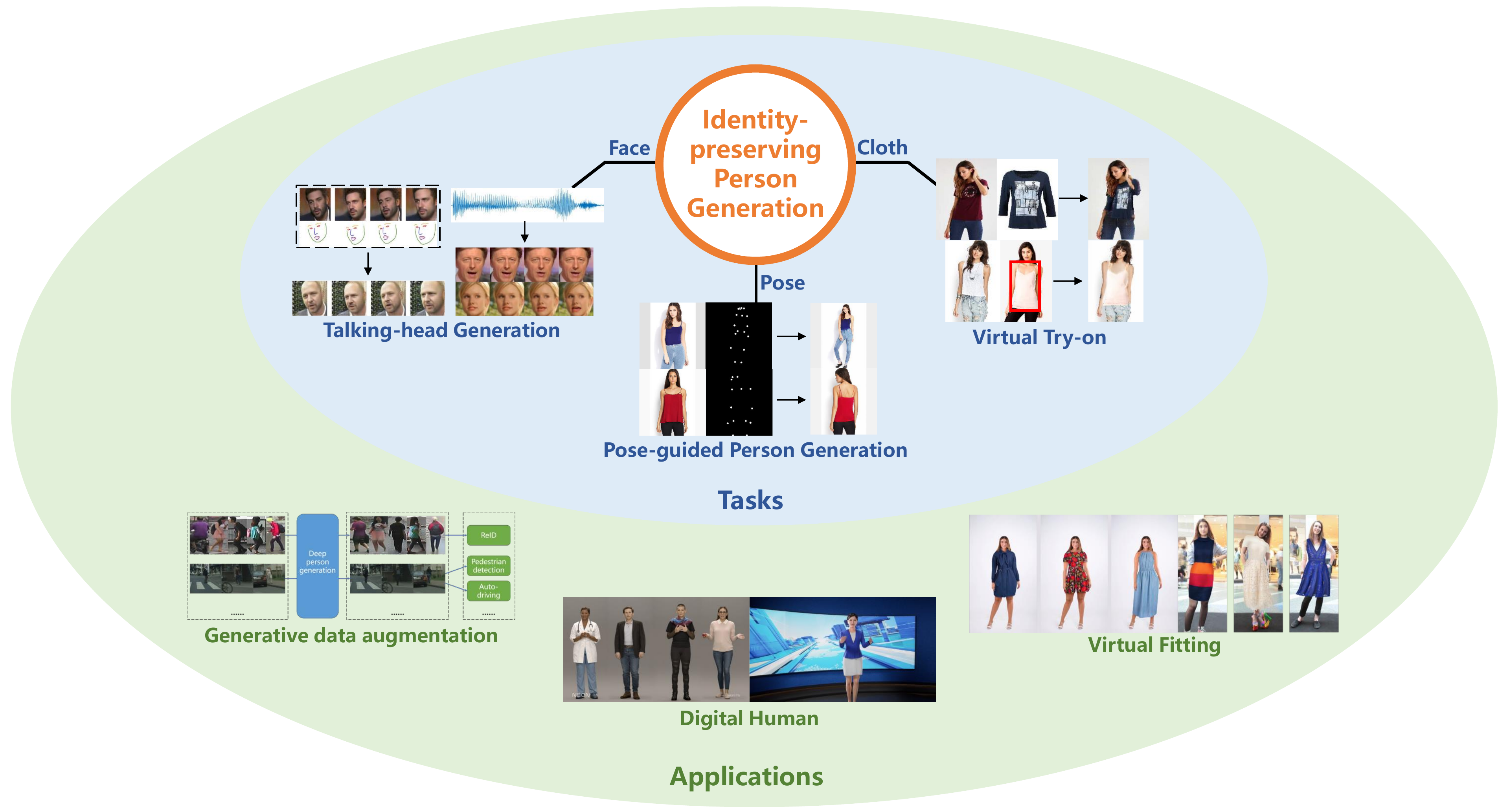}
	\caption{The scope of identity-preserving person generation in this paper. From the three key components (face, pose, garment), we choose three mainstream tasks in recent research, namely talking-head generation, pose-guided person generation and virtual try-on.}
	\label{fig:scope}
\end{figure}

With the advancement of deep learning, people are no longer satisfied with the visual understanding of camera-taken photos/videos. Visual content generation emerges as another research direction since images and videos are much more efficient for information presentation and exchange. Person generation is to synthesize person images and videos as realistically as possible. The long-term goal is to generate digital humans with life-like appearances, expressions and behaviors as real persons. As an emerging area, person generation has attracted lots of research attention, due to its wide applications on digital human\footnote{https://www.neon.life/}, customer service\footnote{https://digitalhumans.com/}, telepresence, art/fashion design and Metaverse.

\begin{figure}[t]
	\centering
	\includegraphics[width=0.99\textwidth]{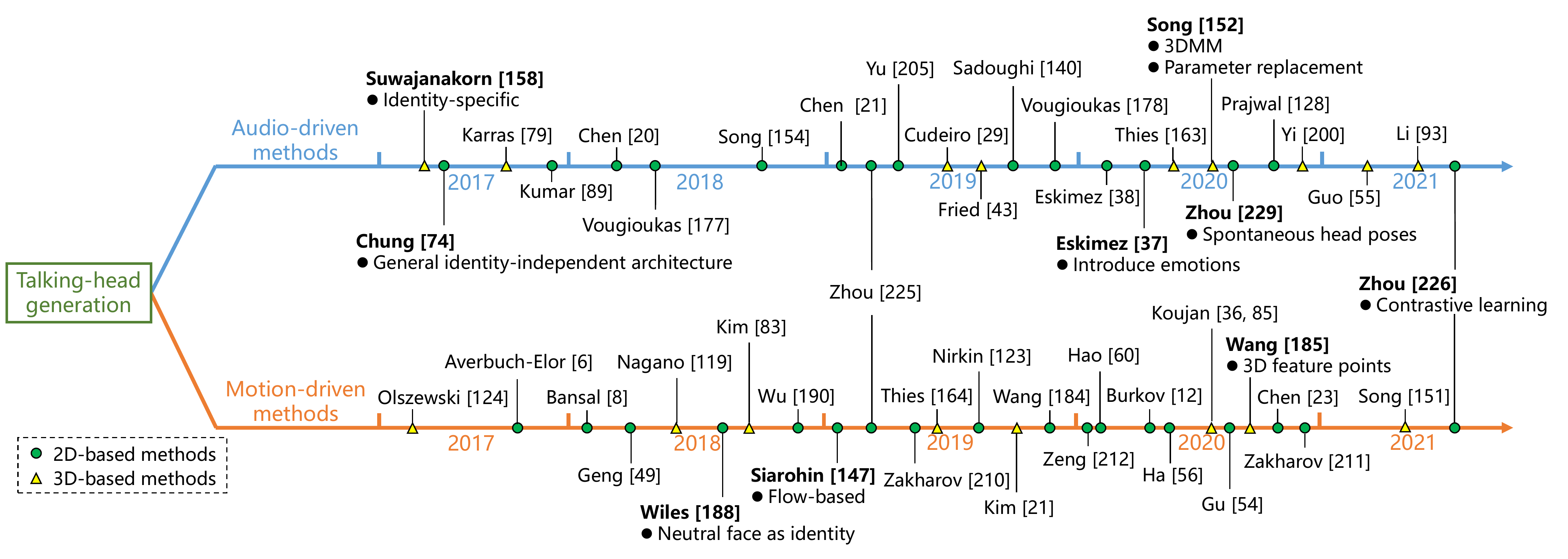}
	\caption{An overview of recent works on talking-head generation. Along the time axis, the number of works is increasing rapidly. For audio-driven generation, the trending technique is mixed with 2D and 3D methods. For motion-driven generation, 3D methods are developed earlier, but 2D methods are surging recently, due to the advancement of GAN (Generative Adversarial Network) \cite{GAN}.}
	\label{Talking-head distribution}
\end{figure}

\begin{figure}[ht]
	\centering
	\includegraphics[width=0.99\textwidth]{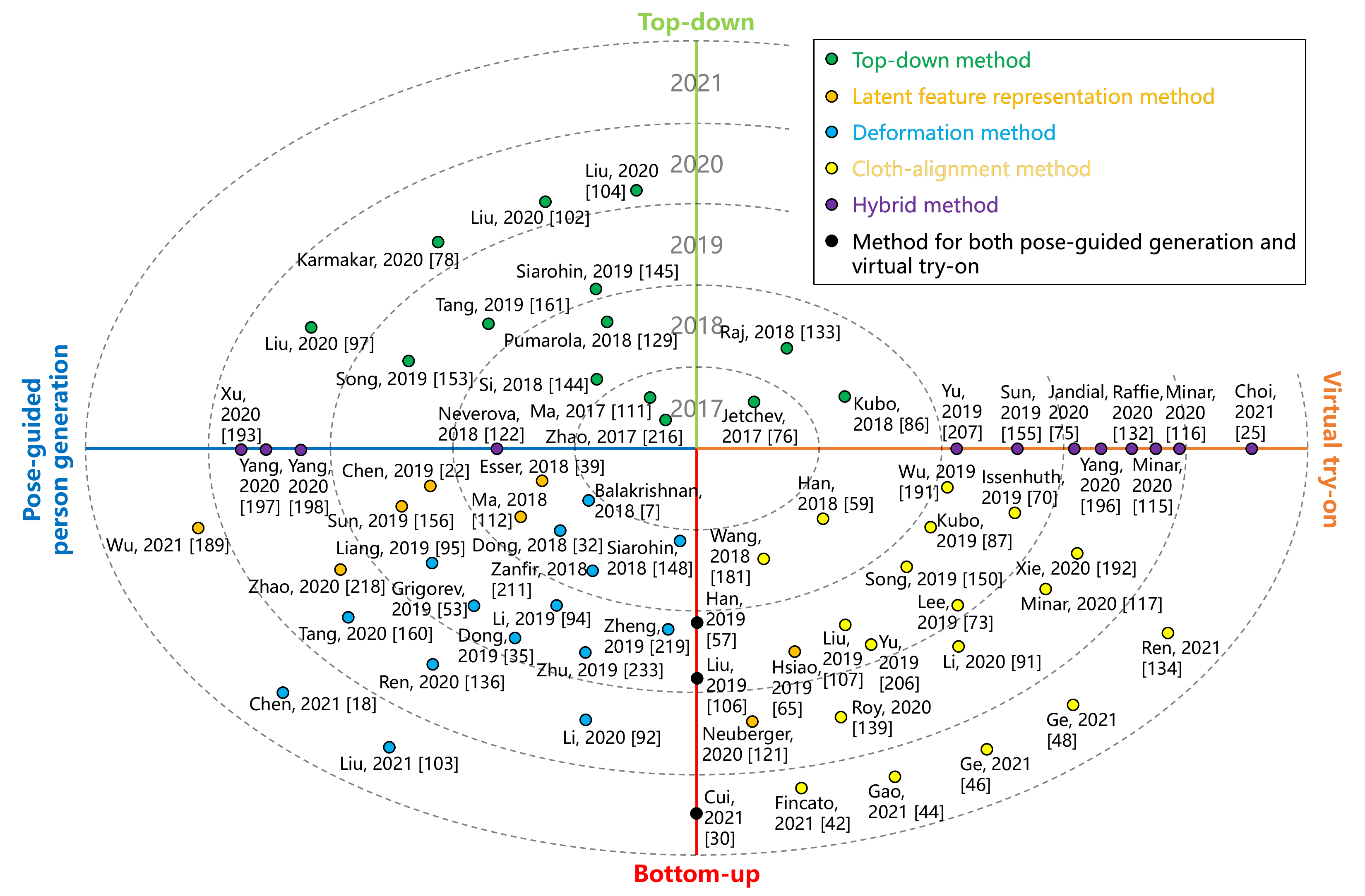}
	\caption{An overview of recent works on pose-guided and garment-oriented person generation since 2017. The left and right halves represent pose and garment transfer, respectively. The upper and lower halves denote top-down and bottom-up methods, respectively. The concentric dotted ellipses (inner to outer) are the time axis.}
	\label{HPT HGT distribution}
\end{figure}

For person images, identity, motion and appearance information are three important components. Identity refers to attributes used to recognize the identity of a person, including but not limited to the face, body shape and height (gait). Among them, face serves as the most commonly adopted media. Motion refers to the body gestures and facial expressions conveyed by a person. Appearance refers to the garments and accessories worn by a person. In some cases, non-identity information such as hairstyle and beard style can also be included as part of appearance information. Generally, in many practical applications such as film shooting, fashion performance and social robots, the demand for motion and appearance generation are highly favored. In this survey, we particularly focus on the identity-preserving person generation, where new face/body dynamics are synthesized while preserving the identity information. Importantly, this task is essential to emerging applications such as digital human, social robots.

Generally speaking, there are three key components (face, pose, cloth) to synthesize a person image or video. We choose the mainstream task in recent research from each component, as shown in Fig.~\ref{fig:scope}. 
\emph{Face generation} is the most popular area and has been extensively studied for years. Facial GANs (Generative Adversarial Networks), including unconditional face generation, facial attributes editing and Deepfakes, are already well reviewed in previous paper \cite{DeepFakeSurvey, DeepFake, MediaForensics&DeepFakes, FaceGenerationSurvey}. Meanwhile, with the easily accessible tools (e.g., DeepFaceLab \cite{DeepFaceLab}), forged facial videos also pose severe social and ethical problems, such as the threats of Deepfakes in the Presidential election. Recently, the identity-preserving facial generation quickly gains popularity, and the most representative branch is the talking-head generation. It has also drawn much attention due to various applications such as video conferencing and customer service. 
\emph{Pose-guided person generation}\footnote{Also known as ``Pose Transfer'' or ``Gesture-to-Gesture Translation'' in other literature.} is another popular task, where a new person image is generated given a conditioning pose. This task is essential to a number of motion-aware generation tasks, such as dancing and sports synthesis. 
\emph{Garment-oriented generation}\footnote{Also known as ``Garment Transfer'' or ``Appearance Transfer'' in some works.} is to synthesize new clothes based on conditioning inputs. It is fundamentally important for virtual try-on and other garment manipulation tasks, i.e., text-guided garment manipulation, garment inpainting.

Fig.~\ref{Talking-head distribution} shows the literature map for talking-head generation. Along the time axis, the number of works has increased sharply in recent years. Roughly two branches of techniques are developed, depending on the driving signals, i.e., audio or motion. For both directions, 2D solutions are developed relatively earlier, and thus draw more research attention so far. Meanwhile, 3D-based methods, as an important complement, are also extensively explored recently. More details are covered in Section~\ref{talking head}.

For pose and garment generation, their technical routines roughly follow a similar pattern. Fig.~\ref{HPT HGT distribution} plots an overview of existing literature. Overall, the following observations can be clearly identified.
(1) The number of works surged rapidly over years (see the time axis). (2) In general, more bottom-up methods are favored over top-down ones. (3) For both pose-oriented and garment-oriented generation, the trend is first from top-down to bottom-up, and then to hybrid methods. (4) The most popular methods for pose and garment transfer are the deformation methods and cloth-alignment methods, respectively. Details are discussed in Section~\ref{PosePG} (pose) and Section~\ref{GarmentPG} (garment).

Despite the rapid development and rich literature, there is no systematic survey for identity-preserving person generation. In this paper, we comprehensively review techniques of identity-preserving person generation in terms of face, pose and garment synthesis.  
Similar works are either focusing on a small sub-area or addressing the topic from a different angle.
Liu \textit{et al}. \cite{GANforImgVid} focus on generic image and video synthesis, while Kammoun \textit{et al}. \cite{FaceGenerationSurvey} highlight face-oriented generation and reenactment. Ruben \textit{et al}. \cite{DeepFakeSurvey} review facial attribute editing and deepfake techniques. Cheng \textit{et al}. \cite{FashionSurvey} conduct a fashion-related survey covering style transfer and pose transformation. Ghodhbani \textit{et al}. \cite{VirtualTryOnSurvey} review the image-based virtual try-on from the angle of fashion. However, these works are all with different focuses in partially overlapping domains, and there is no systematic survey with a broad but focused view on identity-preserving person generation and its applications. Generally, our work has the following contributions:

\begin{itemize}
  \item We provide a systematic survey of identity-preserving person generation from the face, pose, and garment synthesis. To the best of our knowledge, this is the first review of person generation of this kind.
  \item From the three key components (face, pose, garment), we choose three mainstream tasks in recent research, namely \textit{talking-head generation}, \textit{pose-guided person generation} and \textit{virtual try-on}. We provide a comprehensive and in-depth review of state-of-the-art methods. Meanwhile, we summarize the common points of these three tasks from the viewpoint of decomposition.
  \item Mainstream benchmarks and metrics are summarized. Meanwhile, we summarize the main applications: generative data augmentation, virtual fitting and digital human.
  \item We list some possible future directions, to inspire researchers related to person generation.
\end{itemize}

The remaining parts of this survey are organized as follows. Section~\ref{talking head} reviews talking-head generation driven by audio or motion inputs. Section~\ref{PosePG} discusses pose-guided person image/video generation. Section~\ref{GarmentPG} summarizes garment-oriented person generation, namely virtual try-on and other garment manipulation tasks. Section~\ref{Datasets and Metrics} summarizes popular benchmarks and metrics used in person generation. Section~\ref{Common Discussion} concludes the common points of three tasks. Section~\ref{Applications} illustrates major applications of deep person generation. Section~\ref{Future} discusses possible future directions worth further exploration.
\section{Talking-head Generation}
\label{talking head}

Talking-head generation aims to synthesize a person talking image or video, driven by motion, audio or text, which is an important branch of dynamic face generation. As a basis for the subsequent content, we briefly summarise the classic methods on facial image generation.

Goodfellow \textit{et al.} \cite{GAN} propose the Generative Adversarial Network (GAN), where the idea of game adversarial is adopted in network training. Deep Convolutional GAN (DCGAN) \cite{DCGAN} introduces deep convolutions in GAN, which stabilizes the training process. These methods can be directly adopted for unconditional face generation. 
For conditional face generation, Conditional GAN (CGAN) \cite{CGAN} introduces additional inputs to guide the generator for image-to-image translation \cite{FaRGAN, FSGAN} and face reenactment \cite{ReenactGAN, WarpGuidedGAN}. Furthermore, controllable GANs are introduced to control specific attributes of faces. For example, StyleGAN \cite{StyleGAN} uses style codes to control the overall style of the facial image.
These methods serve as the foundation for the subsequent talking-head generation.

\begin{figure}[t]
	\centering
	\includegraphics[width=0.8\textwidth]{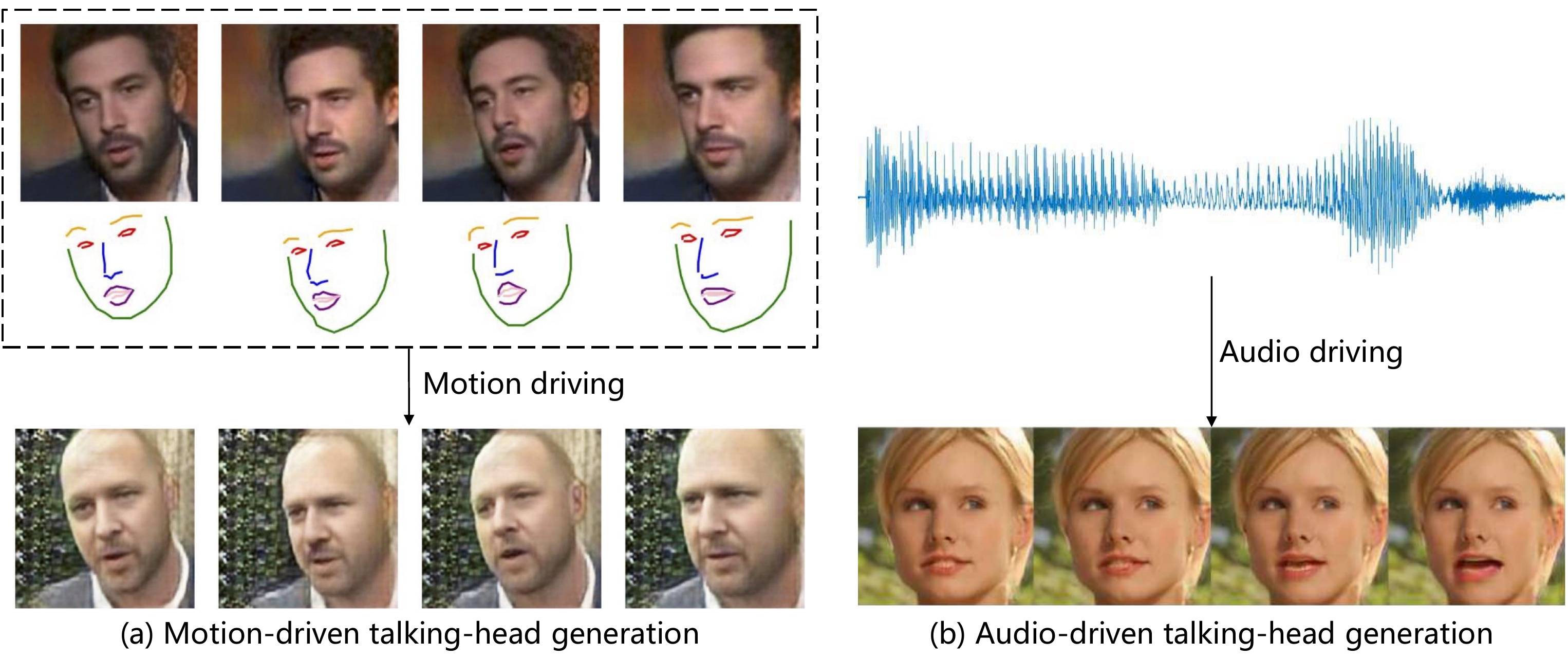}
	\caption{Illustration of motion-driven (a) and audio-driven (b) talking-head generation.}
	\label{audio and motion}
\end{figure}

Talking-head generation is crucial for several applications, including video conferencing, virtual anchors and customer services. As shown in Fig.~\ref{audio and motion}, these methods can generally be divided into two categories, motion-driven and audio-driven, depending on the driving signal. Note that the ``text driven'' (text-to-video) generation is a natural extension of ``audio-driven'' (speech-to-video) since the only difference is the well-developed text-to-speech technique \cite{ObamaNet, WriteASpeaker}. 
Therefore in this survey, we consider the text-driven branch as a variant of the audio-driven problem. Tab.~\ref{Talking head summary} summarizes the representative works on talking-head generation.

\begin{table}[ht]
	\centering
	\caption{Summary of talking-head generation methods. ID D / I: the method is ``Identity-dependent'' or ``Identity-independent''. D: Dependent. I: Independent. H: Hybrid. 3D model: the method use the 3D head model or not. Change pose: the method changes the head poses of source images/videos or not.}
	\scalebox{0.65}{
	\begin{tabular}{p{2.9cm}|p{7cm}|c|c|c|c}
		\hline
		References & Key idea & Driving factor & ID D/I & 3D model & Change pose \\
		\hline
		\hline
		Bansal \cite{RecycleGAN} & Cycle-consistency video-to-video generation & Video & D & $\times$ & $\sqrt{}$ \\
		\hline
        \cite{Portrait2Life, WarpGuidedGAN} & Landmarks-guided warping and detail refinement & Image or video & I & $\times$ & $\sqrt{}$ \\
		\hline
		Wu \cite{ReenactGAN} & Video to landmarks to video & Video & H & $\times$ & $\sqrt{}$ \\
		\hline
		Zakharov \cite{FewShotTalkingHead, FastBiLayerNSofOSRHA} & Landmark-driven few-shot adversarial learning & Image or video & I & $\times$ & $\sqrt{}$ \\
		\hline
		Gu \cite{FLNet} & Landmark-driven GAN with warping and appearance streams & Image or video & I & $\times$ & $\sqrt{}$ \\
		\hline
		\cite{FaRGAN, FSGAN} & Landmark-driven GAN & Image or video & I & $\times$ & $\sqrt{}$ \\
		\hline
		Wang \cite{FewShotVid2vid} & Landmark-driven few-shot video-to-video & Video & I & $\times$ & $\sqrt{}$ \\
		\hline
		Chen \cite{PuppeteerGAN} & Landmarks to semantic map to result & Image or video & I & $\times$ & $\sqrt{}$ \\
		\hline
		Tripathy \cite{IcFace} & Action units based face reenactment & Image or video & I & $\times$ & $\sqrt{}$ \\
		\hline
		Tripathy \cite{FACEGAN} & Action units based landmarks transformer & Image or video & I & $\times$ & $\sqrt{}$ \\
		\hline
		\cite{X2Face, DAEGAN, NHRwithLPD, MarioNETte} & Identity and pose features extraction and fusion & Image or video & I & $\times$ & $\sqrt{}$ \\
		\hline
		Siarohin \cite{FOMMforIA} & Flow warping based video-to-video generation & Video & I & $\times$ & $\sqrt{}$ \\
		\hline
		\cite{RDFT, paGAN} & 3D reconstruction based GAN & Video & I & $\sqrt{}$ & $\times$ \\
		\hline
		Thies \cite{DNR} & 3D reconstruction and parameter replacement & Video & I & $\sqrt{}$ & $\times$ \\
		\hline
		Kim \cite{DVP} & 3D reconstruction and parameter replacement & Video & I & $\sqrt{}$ & $\sqrt{}$ \\
		\hline
		Kim \cite{StylePreservingVD} & 3D reconstruction and parameter replacement & Video & I & $\sqrt{}$ & $\times$ \\
		\hline
		Koujan \cite{Head2Head, Head2Headpp} & 3D reconstruction and parameter replacement & Video & I & $\sqrt{}$ & $\sqrt{}$ \\
		\hline
		\cite{OneShotFVNTHS, EverythingTalking} & 3D Keypoints extraction and flow warping & Video & I & $\sqrt{}$ & $\sqrt{}$ \\
		\hline
		\cite{Speech2Vid, LMG} & Audio and identity features extraction and fusion & Audio & I & $\times$ & $\times$ \\
		\hline
		Zhou \cite{TFGbyADAVR} & Person and word features extraction and fusion & Video or audio & I & $\times$ & $\times$ \\
		\hline
		Zhou \cite{PCAVS} & Contrastive-learning based feature extraction & Video and audio & I & $\times$ & $\times$ \\
		\hline
		Chen \cite{ATVGnet} & Audio to landmarks to video & Audio & I & $\times$ & $\times$ \\
		\hline
		Yu \cite{Face2Vid} & Audio \& text to landmarks to video & Audio and text & I & $\times$ & $\times$ \\
		\hline
		Vougioukas \cite{E2ESDFAwithTGAN} & Aduio-driven GAN & Audio & I & $\times$ & $\times$ \\
		\hline
		\cite{TFGbyCRAN, RSDFA} & RNN-based GAN with three discriminators & Audio & I & $\times$ & $\times$ \\
		\hline
		\cite{CSG, SDTFGfromEmotion} & Audio \& emotion driven GAN & Audio & I & $\times$ & $\times$ \\
		\hline
		Eskimez \cite{E2ETFfromNoisySpeech} & RNN-based GAN & Audio & I & $\times$ & $\times$ \\
		\hline
		Zhou \cite{MakeltTalk} & Audio-driven landmark prediction & Audio & I & $\times$ & $\sqrt{}$ \\
		\hline
		Prajwal \cite{LipSyncExpert} & Wav2Lip: GAN + pre-trained lip-sync expert & Audio & I & $\times$ & $\sqrt{}$ \\
		\hline
		Kumar \cite{ObamaNet} & Text to audio to keypoints to video & Text & D & $\times$ & $\times$ \\
		\hline
		Suwajanakorn \cite{SynthesisObama} & Audio to shapes to mouth images to video & Audio & D & $\sqrt{}$ & $\times$ \\
		\hline
		Karras \cite{ADFAbyPoseEmotion} & Audio \& emotion to 3D model & Audio & D & $\sqrt{}$ & $\times$ \\
		\hline
		Cudeiro \cite{VOCA} & VOCA: speech to 3D model network & Audio & I & $\sqrt{}$ & $\times$ \\
		\hline
		Thies \cite{NVP} & Audio to expressions to 3D model to video & Audio & H & $\sqrt{}$ & $\times$ \\
		\hline
		Song \cite{EverybodyTalk} & 3D reconstruction and parameter replacement & Audio & I & $\sqrt{}$ & $\times$ \\
		\hline
		Yi \cite{TFwithPersonalizedPose} & 3D reconstruction and parameter replacement & Audio & I & $\sqrt{}$ & $\sqrt{}$ \\
		\hline
		Guo \cite{ADNeRF} & Audio-driven neural radiance fields & Audio & D & $\sqrt{}$ & $\sqrt{}$ \\
		\hline
		Fried \cite{TextBasedTalkingHead} & 3D reconstruction and parameter recombination & Text & D & $\sqrt{}$ & $\times$ \\
		\hline
		Li \cite{WriteASpeaker} & Text-driven 3D parameter generation & Text & D & $\sqrt{}$ & $\sqrt{}$ \\
		\hline
	\end{tabular}}
	\label{Talking head summary}
\end{table}

\subsection{Motion-driven Talking-Head Generation}

Motion-driven branch adopts the driving factor of motions, in which the motion includes two parts: head pose and facial expression. This task is also known as ``face reenactment'', which usually manipulates head pose and facial expression simultaneously. There are also methods to manipulate only facial expressions, known as ``Expression Swap''. Motion-driven talking-head generation has many applications in telepresence (e.g., video conferencing, multiplayer online games). Technically, two lines of research can be identified: 2D-based and 3D-based methods, depending on their internal representation of the head model.

\subsubsection{2D-based Methods}

Existing 2D-based works can be roughly grouped into three categories, depending on their intermediate facial representation, i.e., facial landmarks, latent features and action units.

Facial landmark driving methods \cite{FewShotTalkingHead, FLNet, FaRGAN, PuppeteerGAN, WarpGuidedGAN, ReenactGAN, Portrait2Life, FSGAN, FastBiLayerNSofOSRHA, FewShotVid2vid, FACEGAN, FOMMforIA} adopt explicit facial landmarks to encode motions. 
Averbuch-Elor \textit{et al.} \cite{Portrait2Life} and Geng \textit{et al.} \cite{WarpGuidedGAN} warp faces based on landmarks for a coarse result, and adopt a generative adversarial network for refinement. Furthermore, Gu \textit{et al.} \cite{FLNet}, Hao \textit{et al.} \cite{FaRGAN} and Nirkin \textit{et al.} \cite{FSGAN} warp faces on the feature-map level, also based on landmarks. Chen \textit{et al.} \cite{PuppeteerGAN} introduce facial semantic maps to further improve the visual quality. Siarohin \textit{et al.} \cite{FOMMforIA} estimate flow between the source and driving frames, and then warp source image to target frames based on these flows.
However, facial landmarks contain additional identity information. Some works improve the methods based on this point. Zakharov \textit{et al.} \cite{FewShotTalkingHead, FastBiLayerNSofOSRHA} directly map facial landmarks to images, modulated by the identity features. Besides, their few-shot adversarial learning enables training with even fewer images. Wu \textit{et al.} \cite{ReenactGAN} adopt a person-specific transformer to warp landmarks into a specific identity.

Latent feature driving methods \cite{X2Face, TFGbyADAVR, DAEGAN, NHRwithLPD, MarioNETte, PCAVS} are to extract the identity feature from the source video and the motion feature from the driving video, and then fuse them together to generate the target video.
The identity feature could be explicit. X2Face \cite{X2Face} adopts a neutral face image as an explicit identity representation and then warps this neutral face with driving motions. Meanwhile, some works extract implicit identity features. Ha \textit{et al.} \cite{MarioNETte} design an attention block for effective extraction of identity features. Zhou \textit{et al.} \cite{PCAVS} use a contrastive learning strategy to decompose the identity feature and non-identity feature (pose and facial movements) from talking-head videos.

Action units are the coding systems for describing facial expressions. Some methods \cite{IcFace, FACEGAN} use action units to represent identity-independent facial motions. Tripathy et al. \cite{IcFace} use the source image and action units extracted from the driving image to generate the result. This method has a good decoupling of identity and motion, but the precision is limited. They also propose FACEGAN \cite{FACEGAN} to transform landmarks using the action units. The transformed landmarks replace the driving landmarks as the intermediate facial representation. This method ensures that the generated results do not carry the driving face identity information. Meanwhile, it controls the motion more accurately.

\subsubsection{3D-based Methods}

Different from the 2D branch, 3D methods are mainly based on 3D face modeling. Vlasic \textit{et al.} \cite{FaceTransfer} and Garrido \textit{et al.} \cite{VDub} use dubber video to guide lip motions based on 3D face models. Thies \textit{et al.} \cite{RealTimeETforFR, Face2Face} adopt video face tracking to extract 3D face models, and then apply expression transfer for lip motion generation. Note that Face2Face \cite{Face2Face} later becomes the prototype for many subsequent 3D methods.

The above methods need to build special 3D face models manually in advance. Recent works \cite{DVP, StylePreservingVD, Head2Head, Head2Headpp, DNR, paGAN, RDFT} are mostly based on monocular 3D reconstruction, following a similar schema as \cite{Face2Face}. First, monocular 3D reconstruction is adopted to obtain face model parameters for both the source and driving videos. Then, these 3D model parameters are combined to generate the target 3D model. Finally, a video rendering module is applied for video generation. Kim \textit{et al.} \cite{DVP, StylePreservingVD} modify the pose, expression, and eye parameters for fully controllable faces, or modify only the expression parameters to preserve styles. Koujan \textit{et al.} \cite{Head2Head, Head2Headpp} preserve the scale parameter of the source video, to guarantee proper sizes of generated faces. 

Different from the above methods, some works \cite{OneShotFVNTHS, EverythingTalking} utilize 3D facial landmarks to control facial motions. Wang \textit{et al.} \cite{OneShotFVNTHS} propose to extract 3D feature points from videos, instead of using 3D reconstruction. Furthermore, Song \textit{et al.} \cite{EverythingTalking} animate illusory faces with control points extracted from talking video frames.

\subsubsection{Facial Representation Comparison}
The methods summarized in previous sections use many intermediate facial representations. There are five main intermediate facial representations: 2D-based facial landmarks, latent features, action units, 3D-based face model parameters and 3D landmarks. These forms have their own advantages and disadvantages. For example, Tripathy \textit{et al}. \cite{FACEGAN} analyze the pros and cons for landmarks and Wang \textit{et al}. \cite{OneShotFVNTHS} for the 3D face model. We systematically summarize these five intermediate facial representations in Tab.~\ref{motion representations}.

\begin{table}[h]
  \centering
  \caption{Comparison of intermediate facial representations}
  \scalebox{0.6}{
  \begin{tabular}{|c|c|p{5cm}|p{6cm}|p{2cm}|}
    \hline
    Facial representation & Demonstration & Advantages & Disadvantages & Reference \\ \hline
    Facial landmarks
    &
    \begin{minipage}[b]{0.25\columnwidth}
		\centering
		\raisebox{-.5\height}{\includegraphics[width=\linewidth]{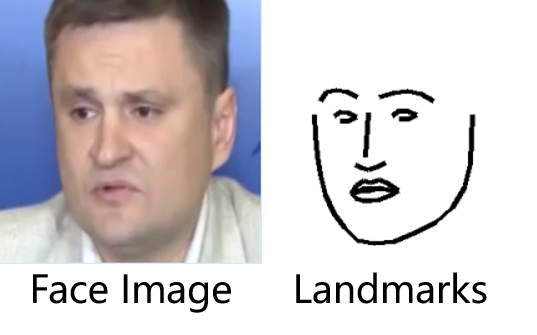}}
	\end{minipage}
    & \makecell[l]{$\bullet$ Strong interpretability and \\ controllability. \\
    $\bullet$ Easy to obtain. \\
    $\bullet$ The training model has high \\ generalization.}
    & \makecell[l]{$\bullet$ There is some identity information as a \\ distraction, including the contour shape of \\ the face. \\ 
    $\bullet$ The pose and expression cannot be \\ decoupled.  \\
    $\bullet$ Poor expression of some details. For \\ example, it is difficult to convey the subtle \\ expression.}
    & \makecell[l]{\cite{FewShotTalkingHead, PuppeteerGAN, Portrait2Life}}
    \\ \hline
    Latent feature
    &
    \begin{minipage}[b]{0.25\columnwidth}
		\centering
		\raisebox{-.5\height}{\includegraphics[width=\linewidth]{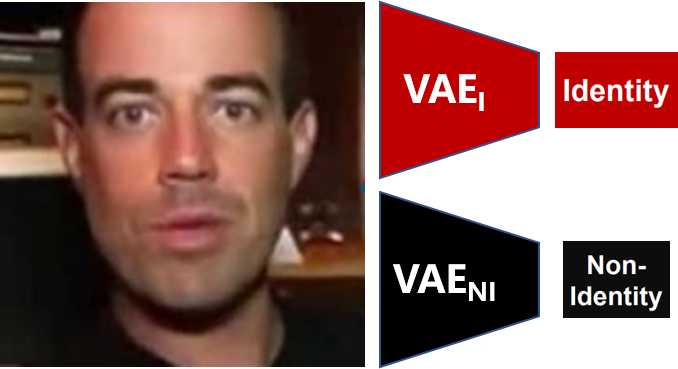}}
	\end{minipage}
    & \makecell[l]{$\bullet$ Have stronger decomposition \\ than facial landmarks. \\
    $\bullet$ Can handle some more delicate \\ decomposition, such as the \\ decomposition between pose and \\  expression.}
    & \makecell[l]{$\bullet$ The interpretability is poor, especially \\ the learning based latent feature. \\
    $\bullet$ It is difficult to ensure that the features \\ have sufficient decomposition. The motion \\ features inevitably contain some identity \\ information. \\
    $\bullet$ The generalization of training model is \\ generally poor. \\
    $\bullet$ Some attribute information will be lost \\ inevitably, resulting in slightly inaccurate \\ or fuzzy results.}
    & \makecell[l]{\cite{X2Face, TFGbyADAVR, PCAVS}}
    \\ \hline
    Action units
    &
    \begin{minipage}[b]{0.25\columnwidth}
		\centering
		\raisebox{-.5\height}{\includegraphics[width=\linewidth]{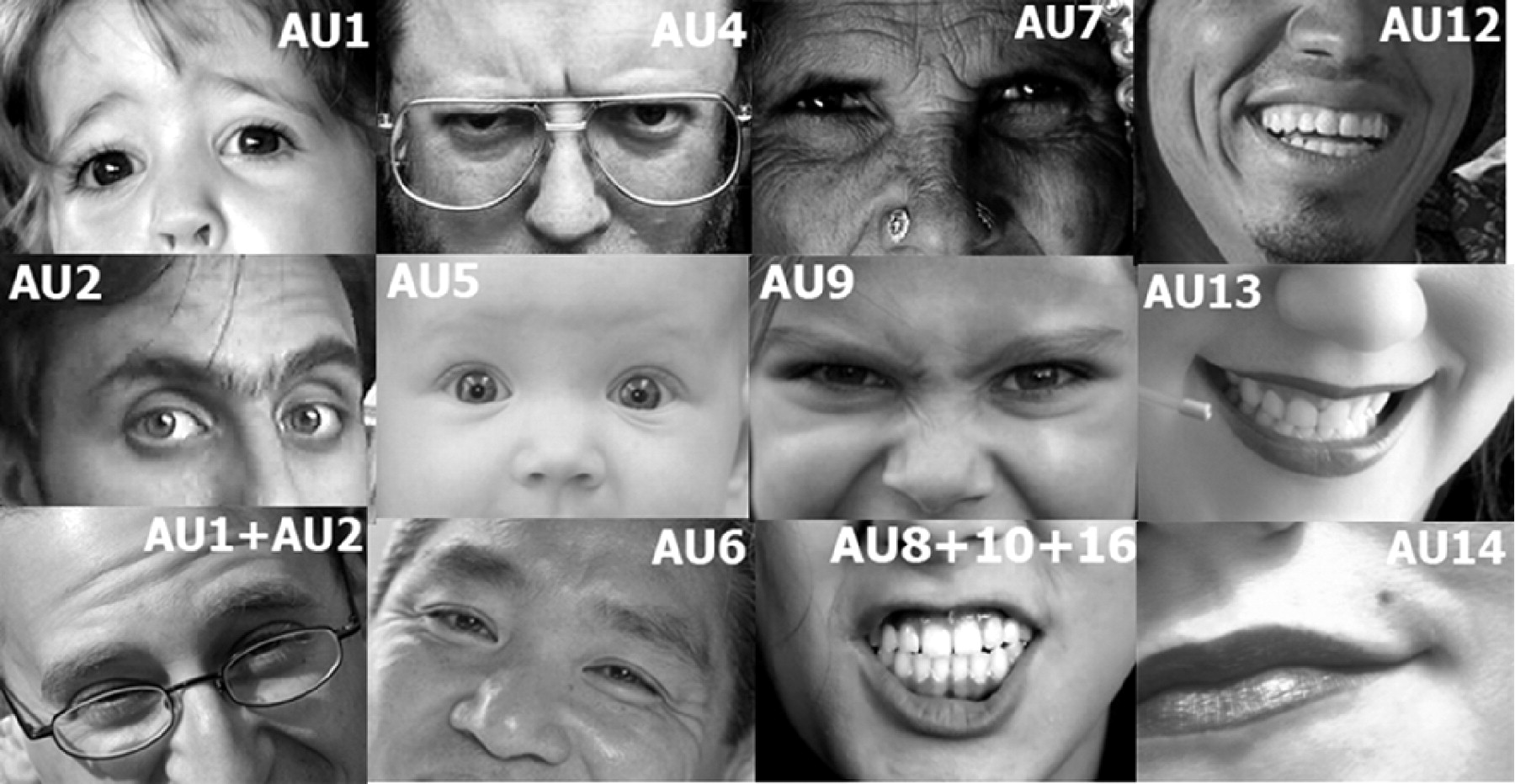}}
	\end{minipage}
    & \makecell[l]{$\bullet$ Have strong decomposition and \\ almost no identity information. \\
    $\bullet$ Have a strong ability to express \\ expression and can express the \\ subtle differences between \\ expressions.}
    & \makecell[l]{$\bullet$ Poor interpretability and controllability \\ making the low quality results.}
    & \makecell[l]{\cite{IcFace, FACEGAN}}
    \\ \hline
    3D model parameters
    &
    \begin{minipage}[b]{0.25\columnwidth}
		\centering
		\raisebox{-.5\height}{\includegraphics[width=\linewidth]{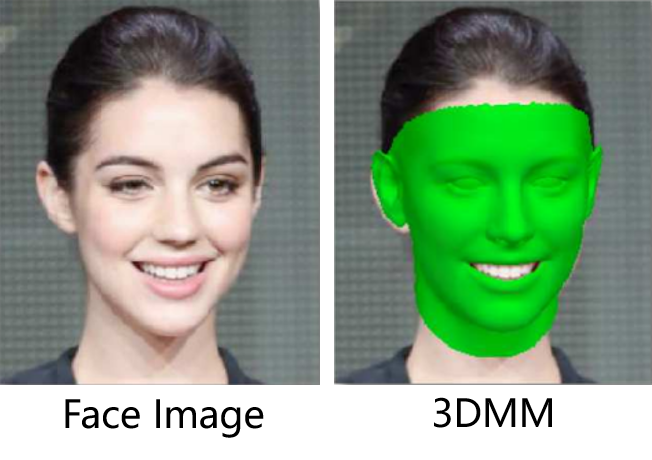}}
	\end{minipage}
    & \makecell[l]{$\bullet$ Strong interpretability and \\ controllability. \\
    $\bullet$ Have strong decomposition. \\ Identity and motion can be well \\ decoupled, and the pose and \\ expression can also be decoupled. \\
    $\bullet$ Results can be generated from \\ almost any view.}
    & \makecell[l]{$\bullet$ It is difficult to obtain 3D parameters. \\ Even with mature monocular 3D \\ reconstruction technology, it takes a \\ certain amount of time to obtain \\ parameters. \\
    $\bullet$ The training model does not have \\ strong generalization and is usually \\ applied to a single identity or a small \\ range of identities. \\
    $\bullet$ Poor ability to express some details, \\ including appearance details and \\ micro-expressions.}
    & \makecell[l]{\cite{DVP, Head2Head, DNR}}
    \\ \hline
    3D landmarks
    &
    \begin{minipage}[b]{0.25\columnwidth}
		\centering
		\raisebox{-.5\height}{\includegraphics[width=\linewidth]{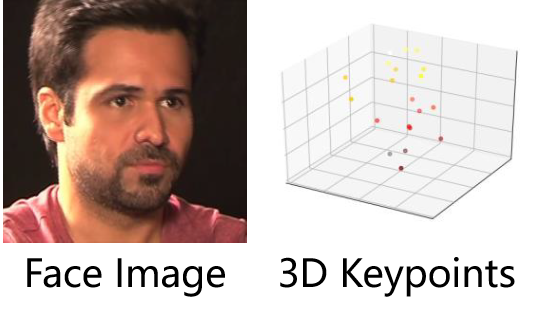}}
	\end{minipage}
    & \makecell[l]{$\bullet$ Relatively easy to obtain. \\
    $\bullet$ Compared with the learning \\ based latent feature, it is more \\ explanatory. \\
    $\bullet$ Strong decomposition, basically \\ not affected by identity \\ information. \\
    $\bullet$ The training model is highly \\ generalized.}
    & \makecell[l]{$\bullet$ Only suitable for local range of views.}
    & \makecell[l]{\cite{OneShotFVNTHS, EverythingTalking}}
    \\ \hline
  \end{tabular}}
  \label{motion representations}
\end{table}

\subsection{Audio-driven Talking-Head Generation}
\label{Audio-driven THG}

Audio-driven talking-head generation is to synthesize a realistic animated video driven by a piece of audio (e.g., speech, singing audio). Recently, this area has been blooming with extensive research attention. Similarly, there are two streams of research, 2D-based and 3D-based methods.

\subsubsection{2D-based Methods}

2D methods mostly adopt landmarks, semantic maps, or other image-like representations during synthesis, which dated back to Bregler \textit{et al}. 1997 \cite{VideoRewrite}. Early works \cite{VideoRewrite, ET2TalkingHead, BiLSTMTalkingHead, TVSA, RFES, VirtualImmortality} mainly use traditional learning methods, such as Hidden Markov Model (HMM) \cite{VideoRewrite, ET2TalkingHead}, LSTM \cite{BiLSTMTalkingHead} and frame retrieval \cite{VirtualImmortality}. Due to the restrictions on method, hardware and data collection, these works apply only to specific identities, and their results are rather preliminary.

Since 2017, GAN-like approaches gradually become popular due to their superior visual quality and strong generalization for identities. Person identities (identity-independent) become popular. Identity-independent methods are mainly divided into two categories, latent feature based and facial landmarks based.

Latent feature based approaches extract audio and identity features with two encoders and fuse them to generate talking heads with new lip motions.
Jamaludin \textit{et al.} \cite{Speech2Vid} propose the first identity-independent architecture by disentangling audio and identity features with two encoders. However, this method does not take into account lip synchronization and video temporal coherence. To solve these problems, Song \textit{et al.} \cite{TFGbyCRAN} and Vougioukas \textit{et al.} \cite{RSDFA} both propose multiple discriminators: frame (image-level fidelity), sequence (video-level fidelity) and syncing (lip reading) discriminators, to improve the visual quality and temporal coherence in video generation.

Facial landmarks based approaches use explicit facial landmarks to connect the audio with lip motions.
Chen \textit{et al.} \cite{ATVGnet} and Yu \textit{et al.} \cite{Face2Vid} utilize facial landmarks as the internal representation to bridge audio and facial images. Furthermore, Sadoughi \textit{et al.} \cite{CSG} and Eskimez \textit{et al.} \cite{SDTFGfromEmotion} introduce emotions as another conditioning input for diverse facial expressions. Zhou \textit{et al.} \cite{MakeltTalk} further apply a speaker-aware animation model to predict spontaneous head poses alongside the audio. To further improve lip synchronization, Prajwal \textit{et al.} \cite{LipSyncExpert} introduce a pre-trained SyncNet as lip-sync discriminator. 

Some methods use text as the driving factor. Kumar \textit{et al.} \cite{ObamaNet} use Char2Wav to transfer textual input into audio. Yu \textit{et al.} \cite{Face2Vid} fuse text and audio together to generate corresponding mouth landmarks.

\subsubsection{3D-based Methods}

Early 3D-based methods pre-build a 3D face model of a specific person, and then render faces based on the 3D model. Compared to 2D solutions, 3D methods are better at motion controlling, especially when synthesizing novel motions and views. But the drawback is the cost of delicate 3D model construction. 
Suwajanakorn \textit{et al.} \cite{SynthesisObama} and Karras \textit{et al.} \cite{ADFAbyPoseEmotion} both adopt a pre-built 3D face model, and then drive the model by learning a sequence (a piece of audio) to sequence (motions of the 3D model) mapping. 

Recent methods tend to directly reconstruct the 3D face model out of the training images or videos. Thies \textit{et al.} \cite{NVP} design an identity-independent Audio2ExpressionNet and an identity-dependent 3D face construction model.
Song \textit{et al.} \cite{EverybodyTalk} replace the 3D expression parameters from the source video with those generated by the input audio. Yi \textit{et al.} \cite{TFwithPersonalizedPose} improve over \cite{EverybodyTalk} by introducing pose parameters for generating natural head motions. To get more refined rendering results, Guo \textit{et al.} \cite{ADNeRF} train two conditional Neural Radiance Fields (NeRFs) \cite{NeRF} to render the head and torso parts, respectively.

Some methods use text instead of audio to drive faces. Fried \textit{et al.} \cite{TextBasedTalkingHead} align phonemes with the 3DMM \cite{3DMM} parameters to generate audio out of input text. Li \textit{et al.} \cite{WriteASpeaker} directly generate 3D head pose, upper face and mouth shape animation parameters according to the input text, skipping the process of generating audio.

\subsection{Discussion}

\begin{figure}[t]
	\centering
	\includegraphics[width=0.78\textwidth]{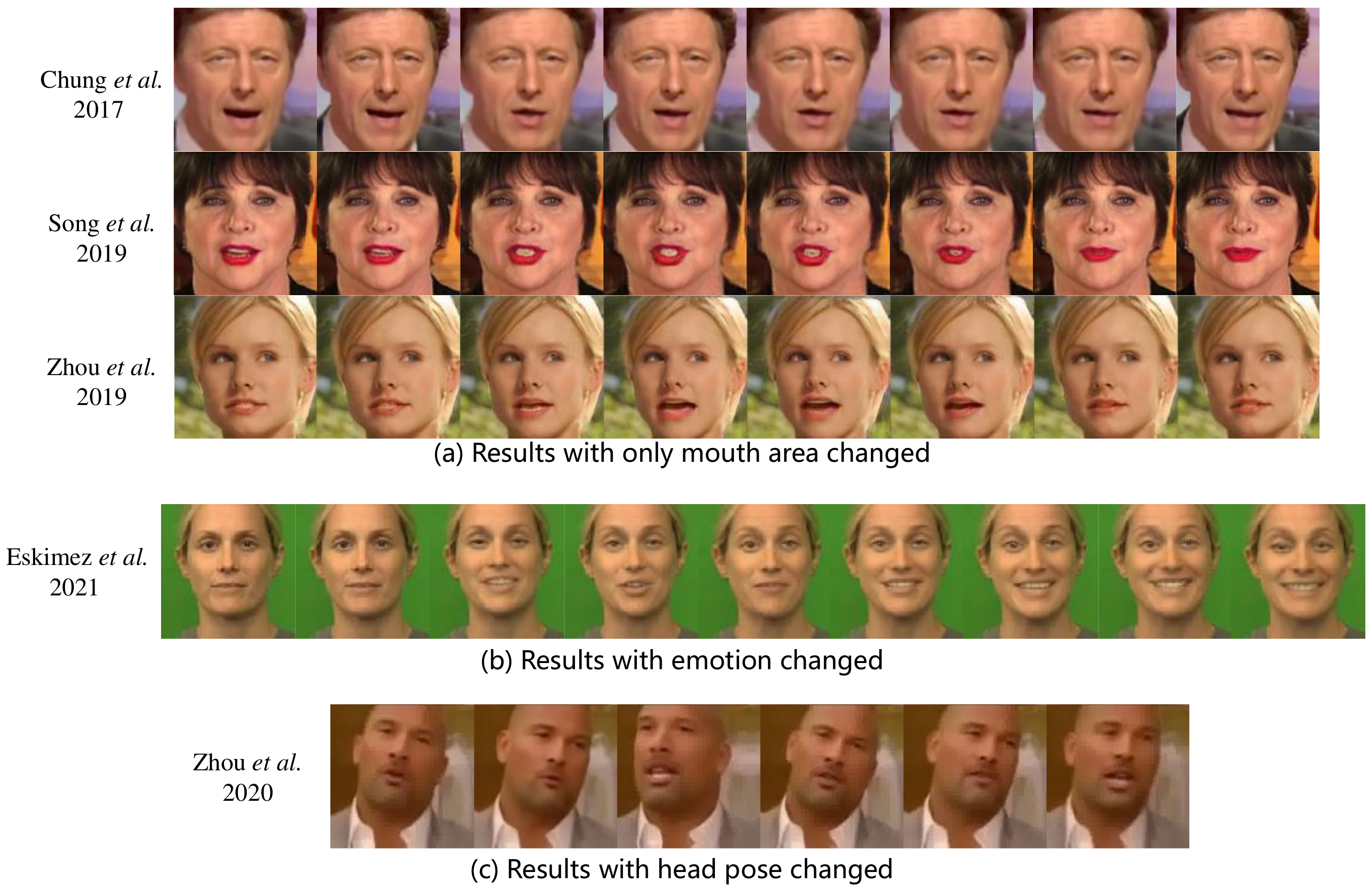}
	\caption{The audio-driven talking-head generation results with (a) only mouth area changed, (b) emotion changed and (c) head pose changed. These results are from Chung \textit{et al.} \cite{Speech2Vid}, Song \textit{et al.} \cite{TFGbyCRAN}, Zhou \textit{et al.} \cite{TFGbyADAVR}, Eskimez \textit{et al.} \cite{SDTFGfromEmotion} and Zhou \textit{et al.} \cite{MakeltTalk}.}
	\label{talking head vivid}
\end{figure}

Facial expressions are becoming vital in recent 2D-based talking-head works. Fig.~\ref{talking head vivid} shows the vividness difference among the results generated by different methods. As is shown in Fig.~\ref{talking head vivid}, early methods only generate and modify the mouth area \cite{VideoRewrite, Speech2Vid, TFGbyCRAN, TFGbyADAVR}, resulting in limited liveliness. Recently, some methods \cite{SDTFGfromEmotion, MakeltTalk} start to explore rich emotions and spontaneous head motions. Especially, Rotger \textit{et al.} \cite{2Dto3DexpressionSwap} bring the 2D facial expression transfer problem into the 3D domain. Based on 3D triangle meshes, they achieved sufficiently detailed expression generation. In the near future, modifying micro-expressions could be possible to appear.

Compared with motion-driven ones, audio-driven tasks are more challenging due to the large domain gap and non-deterministic mapping. It is inherently difficult to establish cross-modal connections among different forms of data: audio, text and video. Furthermore, the same input audio might be interpreted as diverse head motions and facial expressions. In addition, human eyes are highly sensitive to visual artifacts in frames and temporal jitters in videos, making this problem even more challenging.

Either audio- or motion-driven talking-head generation is now leaning towards a solution combining 2D and 3D techniques. In general, GAN-based 2D solutions give more visually plausible results, but 3D solutions are better at motion control. Furthermore, recent advancements in monocular 3D reconstruction also provide convenient 3D models during generation. Regarding performance, this task is still far from mature. For example, generating high-resolution video with micro-expressions is still a major challenge to current methods. 

\section{Pose-guided Person Generation}
\label{PosePG}

\begin{figure}[t]
	\centering
	\includegraphics[width=0.8\textwidth]{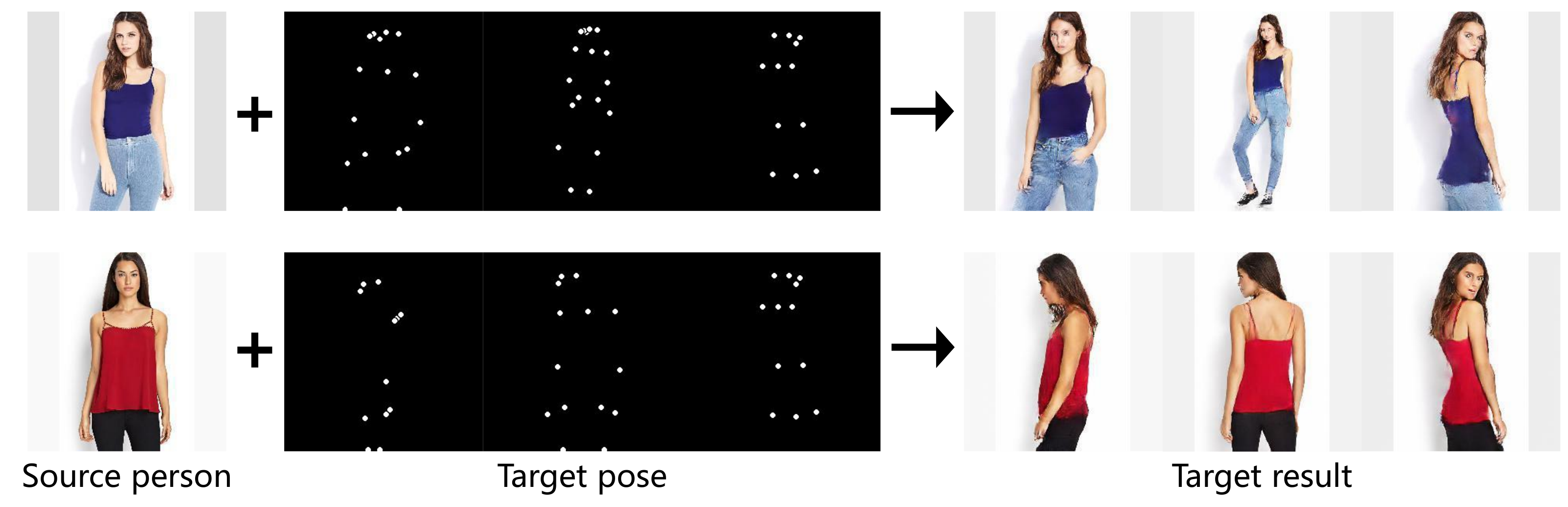}
	\caption{Illustration of pose-guided person generation.}
	\label{HPT}
\end{figure}

Pose-guided person generation aims to generate full-body person images or videos guided by target poses as realistic as possible, as shown in Fig.~\ref{HPT}. The fundamental difference between talking-head is the emphasis on body poses and motions. This task has many potential applications in movie production, online fashion shopping, etc. We focus on two major branches: pose-guided person image and video generation.

\subsection{Pose-guided Person Image Generation}
\label{ssec:pgpg}

\begin{table}[ht]
	\centering
	\caption{Summary of pose-guided person image generation works. Use parsing: use human parsing or not. Separate FG: Separate the foreground of images or not.}
	\scalebox{0.65}{
	\begin{tabular}{p{2.4cm}|p{6.5cm}|c|c|c}
		\hline
		References & Main idea & Use 3D pose & Use parsing & Separate FG \\
		\hline
		\hline
		\multicolumn{5}{c}{Top-down methods} \\
		\hline
		\cite{VariGAN, PG2} & Coarse-to-fine generation & $\times$ & $\times$ & $\times$ \\
		\hline
		Si \cite{Si2018} & Direct generation & $\times$ & $\times$ & $\sqrt{}$ \\
		\hline
		Karmakar \cite{ptGAN} & Direct generation & $\times$ & $\times$ & $\times$ \\
		\hline
		Tang \cite{C2GAN} & Cycle consistency & $\times$ & $\times$ & $\times$ \\
		\hline
		Liu \cite{SemanticAttentionPIG} & Semantic-guided , attention mechanism & $\times$ & $\sqrt{}$ & $\times$ \\
		\hline
		Liu \cite{SegmentationMaskPIG} & Mask-guided generation & $\times$ & $\times$ & $\sqrt{}$ \\
		\hline
		Siarohin \cite{AttentionMHIG} & Multi-source generation, attention mechanism & $\times$ & $\times$ & $\times$ \\
		\hline
		Liu \cite{HighResolutionAT} & High-resolution progressive training & $\times$ & $\times$ & $\sqrt{}$ \\
		\hline
		Pumarola \cite{UPIS} & Cycle consistency & $\times$ & $\times$ & $\times$ \\
		\hline
		Song \cite{UPIGSPT} & Cycle consistency & $\times$ & $\sqrt{}$ & $\times$ \\
		\hline
		\multicolumn{5}{c}{Bottom-up latent feature representation methods} \\
		\hline
		Esser \cite{VUnet} & Feature representation & $\times$ & $\times$ & $\times$ \\
		\hline
		Chen \cite{UPGGAN} & Cycle consistency and feature representation & $\times$ & $\times$ & $\times$ \\
		\hline
		Ma \cite{Disentangle} & Feature representation & $\times$ & $\times$ & $\sqrt{}$ \\
		\hline
		Sun \cite{Sun2019} & Multi-source feature representation & $\times$ &$\times$ & $\times$ \\
		\hline
		Zhao \cite{PoseSequenceHPT} & Pose serialization & $\times$ & $\times$ & $\times$ \\
		\hline
		Wu \cite{DFCNet} & Feature representation & $\times$ & $\times$ & $\times$ \\
		\hline
		\multicolumn{5}{c}{Bottom-up deformation methods} \\
		\hline
		Siarohin \cite{DefGAN} & Local deformation & $\times$ & $\times$ & $\times$ \\
		\hline
		Liang \cite{PCGAN} & Local deformation & $\times$ & $\times$ & $\sqrt{}$ \\
		\hline
		Balakrishnan \cite{Balakrishnan2018} & Local deformation & $\times$ & $\sqrt{}$ & $\sqrt{}$ \\
		\hline
		Dong \cite{WarpingGAN} & Global deformation & $\times$ & $\sqrt{}$ & $\times$ \\
		\hline
		Zheng \cite{UPFL} & Flow warping & $\sqrt{}$ & $\sqrt{}$ & $\sqrt{}$ \\
		\hline
		Han \cite{ClothFlow} & Flow warping & $\times$ & $\sqrt{}$ & $\times$ \\
		\hline
		\cite{HAT, Grigorev2019} & Global deformation & $\sqrt{}$ & $\times$ & $\sqrt{}$ \\
		\hline
		\cite{DIAF, DiOr} & Flow warping & $\times$ & $\times$ & $\sqrt{}$ \\
		\hline
		Liu \cite{LiquidGAN} & Flow warping & $\sqrt{}$ & $\times$ & $\sqrt{}$ \\
		\hline
		Dong \cite{PPGAN} & Decompose human and pose into parts & $\times$ & $\sqrt{}$ & $\times$ \\
		\hline
		\cite{PATN, DIST, XingGAN, PoNA} & Attention mechanism & $\times$ & $\times$ & $\times$ \\
		\hline
		Chen \cite{PMAN} & Multi-attention mechanism & $\times$ & $\times$ & $\times$ \\
		\hline
		Liu \cite{SPAN} & Semantic parsing attention & $\times$ & $\sqrt{}$ & $\times$ \\
		\hline
		\multicolumn{5}{c}{Hybrid methods} \\
		\hline
		Neverova \cite{DPT} & Direct generation and global deformation & $\sqrt{}$ & $\times$ & $\times$ \\
		\hline
		Yang \cite{RATENet, DRN} & Attention mechanism and refinement & $\times$ & $\times$ & $\times$ \\
		\hline
		Xu \cite{NonIconicHPT} & Deformation and refinement in a non-iconic view & $\times$ & $\times$ & $\sqrt{}$ \\
		\hline
	\end{tabular}}
	\label{HPT summary}
\end{table}

\begin{figure}[t]
	\centering
	\includegraphics[width=0.99\textwidth]{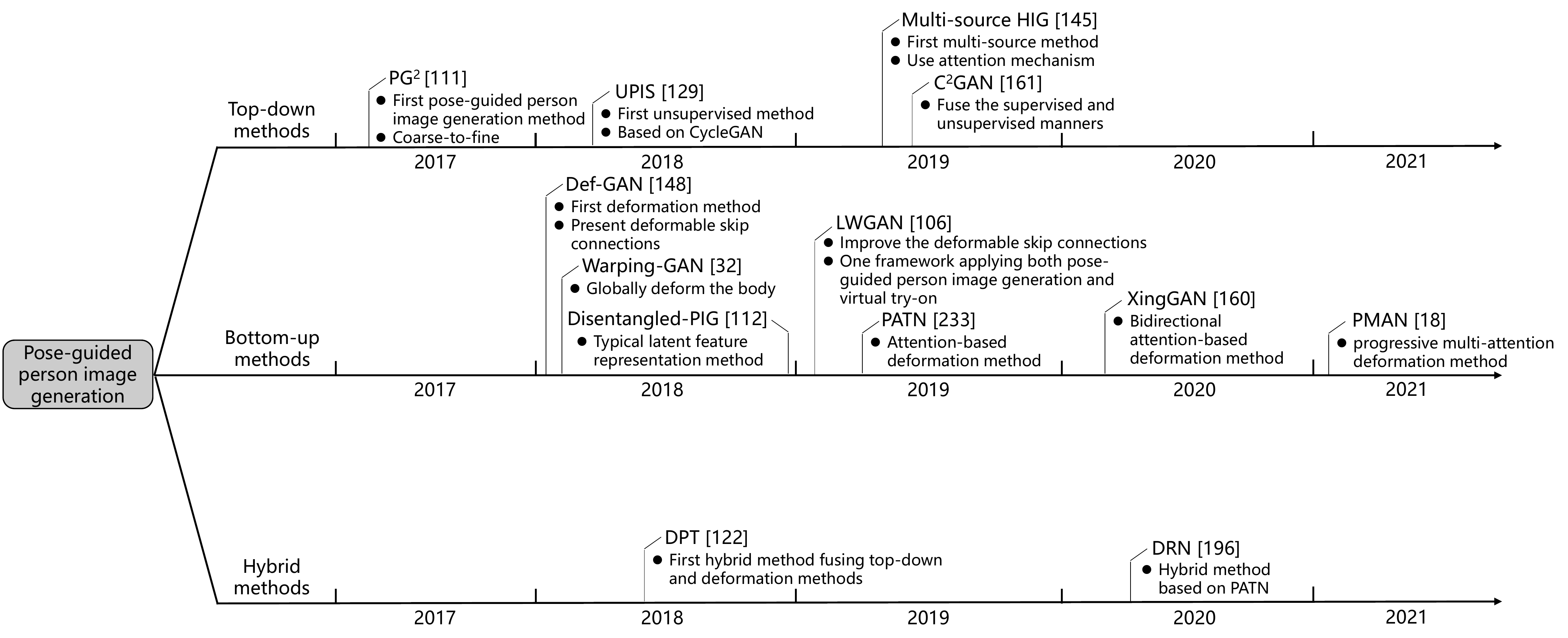}
	\caption{Representative works for pose-guided image person generation.}
	\label{HPT representation}
\end{figure}

\begin{figure}[t]
	\centering
	\includegraphics[width=0.99\textwidth]{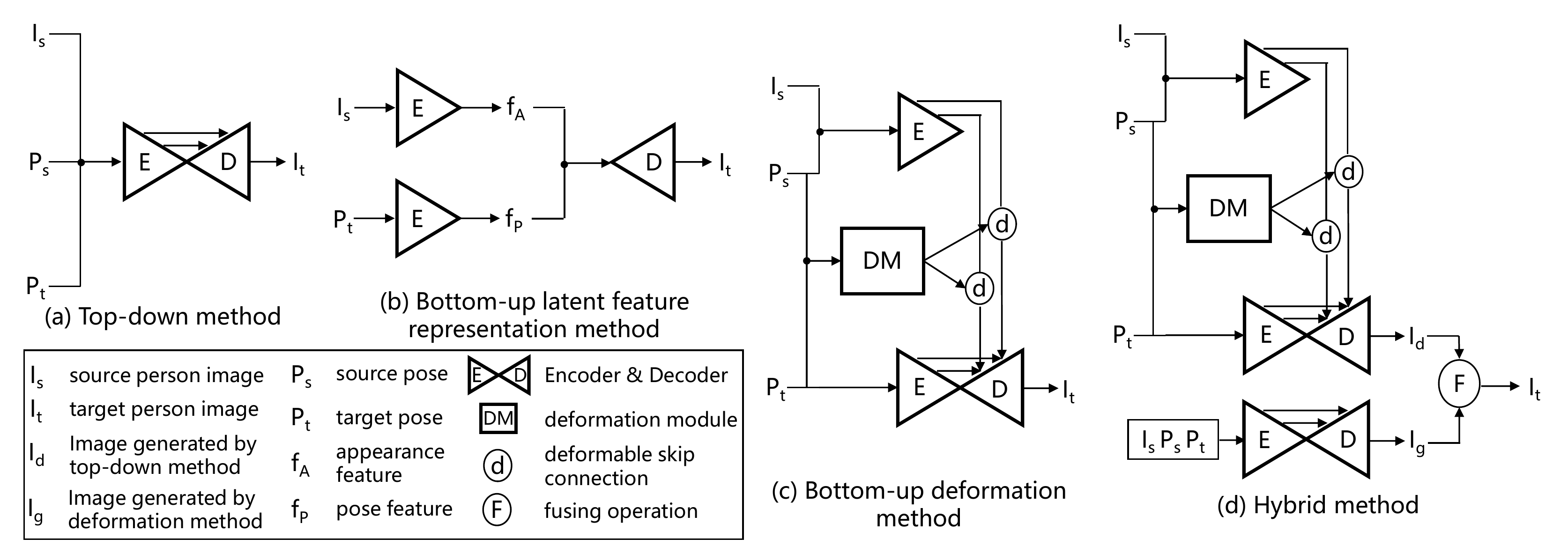}
	\caption{Four typical network structures for pose-guided person image generation. (a) Top-down methods directly concatenate $I_s$, $P_s$ and $P_t$ as input to synthesize $I_t$. (b) Bottom-up latent feature representation methods extract the source appearance feature $f_A$ and target pose feature $f_P$ as internal representations. (c) Bottom-up deformation methods try to deform the feature maps of $I_s$ and $P_s$ into the target pose $P_t$. (d) Hybrid methods fuse the top-down and bottom-up deformation methods into a framework.}
	\label{HPT difference}
\end{figure}

Pose-guided person image generation aims to transform a source person image according to a given target pose. The main challenge lies in the misalignment between the source and target poses. 

The summary on pose-guided person image generation is shown in Tab.~\ref{HPT summary}. Existing methods can be roughly grouped into three categories: top-down, bottom-up and hybrid methods. Concretely, top-down methods directly learn a mapping from the input to desired output image, mostly with a GAN-based network. Bottom-up approaches decompose the whole pipeline into several intermediate components and build up the final result step by step. Hybrid approaches take advantage of both sides. As an overview, Fig.~\ref{HPT representation} shows the representative works from each category, and Fig.~\ref{HPT difference} compares the main network structures.

\subsubsection{Top-down Methods}
\label{HPT Top-down Methods}

Inspired by the GAN-based view synthesis \cite{VariGAN}, Ma \textit{et al}. \cite{PG2} apply conditional GAN \cite{CGAN} for pose-guided person generation, and propose the pioneer top-down solution PG$^{2}$. PG$^{2}$ directly concatenates the source image and target pose as input to synthesize the target image, plus a refinement model to improve the generation details. 

The results generated by PG$^{2}$ still lose many appearance details. For better performance, several improvements are proposed subsequently, including data augmentation \cite{ptGAN} and more informative inputs \cite{SemanticAttentionPIG, SegmentationMaskPIG, AttentionMHIG}. 
Specifically, Siarohin \textit{et al}. \cite{AttentionMHIG} propose a multi-source method, where an attention mechanism is adopted to assign weights to different source images. Liu \textit{et al}. \cite{SegmentationMaskPIG, SemanticAttentionPIG} replace poses with segmentation masks and parsing maps to guide image generation. 

Besides the aforementioned supervised branch, some works explore the unsupervised setting. Pumarola \textit{et al}. \cite{UPIS} introduce the first unsupervised method, which uses the generated target person image with the source pose to regenerate the source image similar to CycleGAN \cite{CycleGAN}. Song \textit{et al}. \cite{UPIGSPT} operates on the semantic maps to bypass the requirement of paired data and generate more appearance details. Tang \textit{et al}. \cite{C2GAN} introduce multiple cyclic losses, i.e., 1 $\times$ image$\rightarrow$image$\rightarrow$image cycle and 2 $\times$ pose$\rightarrow$image$\rightarrow$pose cycles, to introduce more self-supervisions.

Top-down methods can not well preserve the appearance details of the source person, and it is difficult to ensure the identity consistency of the generated results. To solve these problems, bottom-up methods are proposed.

\subsubsection{Bottom-up Methods}
\label{HPT Bottom-up Methods}

Bottom-up methods tend to decompose the whole process into components or steps, where the intermediate results are usually essential for the generation. In general, these methods can be grouped into two categories: latent feature representation and deformation methods.

\textbf{Latent Feature Representation Methods} \cite{VUnet, UPGGAN, Disentangle, Sun2019, PoseSequenceHPT, DFCNet} extracts latent features from the source image and target pose as the intermediate results, to control the generation results. 
The works based on latent feature representation aim at extracting accurate and pure feature information.
A typical way \cite{VUnet, UPGGAN} is to extract appearance features from the source image via VAE \cite{VAE}. Chen \textit{et al}. \cite{UPGGAN} use cycle-consistency \cite{CycleGAN} to support unpaired training data. Zhao \textit{et al}. \cite{PoseSequenceHPT} extract pose features from an interpolated pose sequence, from source pose to target pose. Sun \textit{et al}. \cite{Sun2019} extract appearance features from a set of source images by bidirectional convolutional LSTM. Ma \textit{et al}. \cite{Disentangle} use adversarial training to disentangle the input into three factors: foreground, background and pose.

Compared to top-down methods, latent feature representation methods have better performance and preserve the identity information to a certain extent.

\textbf{Deformation Methods} are another line of research. 
Top-down and latent feature representation methods usually show inferior results for large differences between source and target poses. Deformation methods are proposed to address the above deficiency, by transferring the feature maps of the source person image into the target pose. These methods are better at preserving texture details.
Deformation methods can be further divided into three developmental branches according to different deformation strategies: local deformation, global deformation and attention-based deformation.

\textit{Local deformation methods} \cite{DefGAN, PCGAN, Balakrishnan2018, PPGAN} adopt the strategy of disintegrating the body and partial deformation. 
The first method dates back to Deformable GAN (Def-GAN). Inspired by the spatial transformer networks \cite{STN}, Siarohin \textit{et al}. \cite{DefGAN} propose to locally deform near-rigid body parts at the feature map level. A new Deformable Skip Connection (DSC) is proposed to replace standard skip operation, where the deformed feature and the target pose are concatenated as a comprehensive input.  DSC uses a rough deformation strategy, so some works \cite{PCGAN, Balakrishnan2018} improve over DSC. Especially, Liang \textit{et al}. \cite{PCGAN} further consider the modification of the body shape in the local deformation process.
Besides, a few methods choose not to explicitly deform body parts at the input side, but to deform implicitly with an encoder-decoder structure during generation. Dong \textit{et al}. \cite{PPGAN} introduce Part-Preserving GAN (PP-GAN), which directly takes as input the decomposed body parts and implicitly deforms with a parsing-consistent loss. 
Although the local deformation strategy retains enough appearance details, it ignores the connections between the local parts, resulting in a poor generation effect in these connection areas.

\textit{Global deformation methods} \cite{HAT, WarpingGAN, UPFL, LiquidGAN, ClothFlow, Grigorev2019, DIAF, DiOr} adopt global deformation to warp the human body as a whole. 
Dong \textit{et al}. \cite{WarpingGAN} propose a soft-gated warping-block to predict the transformation grid between parsing maps. 
Some works \cite{UPFL, LiquidGAN, ClothFlow, DIAF, DiOr} estimate the appearance flow from source pose to target pose, to transform the full body. 
Furthermore, 3D models are introduced to improve warping accuracy.
Zanfir \textit{et al.} \cite{HAT} fill in the dense 3D mesh of the source person with textures and then transfer the mesh into the target result. Grigorev \textit{et al}. \cite{Grigorev2019} first extract the 3D DensePose \cite{DensePose} for target pose, and then inpaint the surfaces with a coordinate-based method.
Global deformation methods can preserve enough appearance details, but they are hard to deal with the appearance details of unseen areas.

\textit{Attention-based deformation methods} \cite{PATN, DIST, XingGAN, PoNA, PMAN, SPAN} introduce the attention mechanism for implicit deformation.
Zhu \textit{et al}. \cite{PATN} propose Pose-Attentional Transfer Network (PATN) to assign weights to different image patches according to the pose. 
Ren \textit{et al}. \cite{DIST} generate flow fields to calculate local attention for feature maps extracted from the source image, and thus spatial transformation can be locally operated. 
The above two methods only contain the pose-based attention branch. Furthermore, 
Tang \textit{et al}. \cite{XingGAN} design a novel attention-based network XingGAN including two branches: Shape-guided Appearance-based generation (SA) and Appearance-guided Shape-based generation (AS). A novel crossing connection is introduced to capture the joint attention between image and pose modalities.
Attention-based deformation methods can well predict the appearance details of unseen areas and show great potential.

\subsubsection{Hybrid Methods}
\label{Hybrid Methods}

Top-down methods are better at preserving the overall appearance but not at retaining the local textures, especially there is a large gap between source and target poses. Bottom-up methods deal well with large gaps, but have difficulties in handling occlusions. 
Therefore, the hybrid frameworks are explored to take advantage of both sides.

Neverova \textit{et al}. \cite{DPT} propose the first hybrid method, Dense Pose Transfer (DPT), which fuses the top-down result and deformation output for target image generation. 
Yang \textit{et al}. \cite{RATENet, DRN} combine PATN \cite{PATN} with image refinement \cite{PG2} and apply an alternate updating strategy to facilitate mutual guidance between two modules for better appearance and shape consistency.
Xu \textit{et al}. \cite{NonIconicHPT} propose MR-Net to synthesize person images with distorted poses and cluttered scenes.

\subsection{Pose-guided Person Video Generation}
\label{PGPVG}

It is natural to extend pose-guided generation from image to video domain, by further considering temporal coherence in generated videos. Similar to image generation, we categorize related works into two branches: top-down and bottom-up methods.

\subsubsection{Top-down Methods}

Top-down methods directly map a target pose sequence to a person video via GAN-based networks. The target pose sequence is either given \cite{SAAMG} or predicted by another network \cite{PGHVG, GLTFviaHP, ThePoseKnows, ForecastI2VG}. 
Yang \textit{et al}. \cite{PGHVG} predict the pose sequence via Pose Sequence GAN (PSGAN). 
Walker \textit{et al}. \cite{ThePoseKnows} combine VAE \cite{VAE} with LSTM to learn the distribution of future poses. Meanwhile, pose flow is introduced to improve temporal coherence. 
Furthermore, Villegas \textit{et al}. \cite{GLTFviaHP} propose an analogy generating method to synthesize future frames. Zhao \textit{et al}. \cite{ForecastI2VG} improve over \cite{GLTFviaHP} with motion refinement network for better temporal coherence.

\subsubsection{Bottom-up Methods}

To retain more temporal coherence, bottom-up methods are explored to inject more structured information during video generation.
Wang \textit{et al}. \cite{vid2vid} propose the first video-to-video synthesis approach, where optical flow is adopted to improve temporal coherence.
Chan \textit{et al}. \cite{EverybodyDance} introduce a typical pose-guided video generation method to transfer body motions from another video. A sequence of frames, instead of a single image, is adopted to preserve temporal coherence. Liu \textit{et al}. \cite{ReenactmentHAV} better transfer poses on 3D model.

The above methods \cite{vid2vid, EverybodyDance, ReenactmentHAV} are limited in identity dependency. That is, one has to re-train the model when applying it to a new person. Some researchers attempt to design identity-independent methods \cite{MonkeyNet, FOMMforIA, FewShotVid2vid, DHAVG, DanceDanceGeneartion, DwNet, DSTforPVG, PGPVGinthewild, D2GNet}. 
Siarohin \textit{et al}. \cite{MonkeyNet, FOMMforIA} add a source image as input to generate video for different identities. Furthermore, optical flow between source and driving frames is estimated to preserve temporal coherence.
Wang \textit{et al}. \cite{FewShotVid2vid} also add identify images as additional inputs to control the appearance of the generated video. 
Zhou \textit{et al}. \cite{DanceDanceGeneartion} present a local deformation method that deforms body parts into target pose. 
Ren \textit{et al}. \cite{DSTforPVG} remould image generation network \cite{DIST} to synthesize continuous frames, and introduce a Motion Extraction Network to improve temporal consistency.
Yoon \textit{et al}. \cite{PGPVGinthewild} use the 3D human model to transfer the pose in a temporally consistent way, which can be applied to in-the-wild images.

\subsection{Discussion}


Pose-guided person generation is challenging mainly due to the diversity and complexity of human poses. For large gaps between source and target poses, unseen areas are extremely difficult to synthesize. Top-down methods adopt GAN-based networks to hallucinate unseen areas, but sacrifice the texture details. Bottom-up methods deform the source image into the target pose, which better preserves texture details but is sensitive to occluded areas. 

So far, most pose-guided generation methods belong to the bottom-up branch, while top-down approaches are also evolving fast in visual quality.
Among bottom-up methods, latent feature representation methods are pure implicit learning methods with poor interpretation, but they have a strong ability to predict invisible regions according to visible regions. On the contrary, explicit deformation methods use many strong explanatory strategies including disentangled affine transformation, appearance flow warping and 3D-based deformation, But it is difficult to predict the details of the appearance of invisible areas. While, deformation methods based on the attention mechanism, which implicitly deform the pose in latent feature space, take the advantage of both latent feature representation and explicit deformation methods. They can predict the details of the visible area well and have shown great potential in further improving the texture details. However, large differences between poses, e.g., front to back, are still difficult to handle for now.
The hybrid methods have more potentials in the future in addressing the difficulties from occlusions and texture details. 

Recently, monocular 3D human reconstruction has been developed gradually \cite{PIFu, PIFuHD, PaMIR}. However, only reconstruction based on the full-body image is supported. For pose-guided person generation, many source images do not show complete bodies, so it is difficult to apply the 3D reconstruction idea. 3D models have strong controllability. 3D models have great potential in future pose-guided person generation.
\section{Garment-Oriented Person Generation} 
\label{GarmentPG}

Garment-oriented generation concentrates on generating new clothing of a person image, which has wide applications in the fashion domain. In this section, we focus on two mainstream tasks: virtual try-on and garment manipulation.

\subsection{Virtual try-on}
\label{HGT}

\begin{figure}[t]
	\centering
	\includegraphics[width=0.7\textwidth]{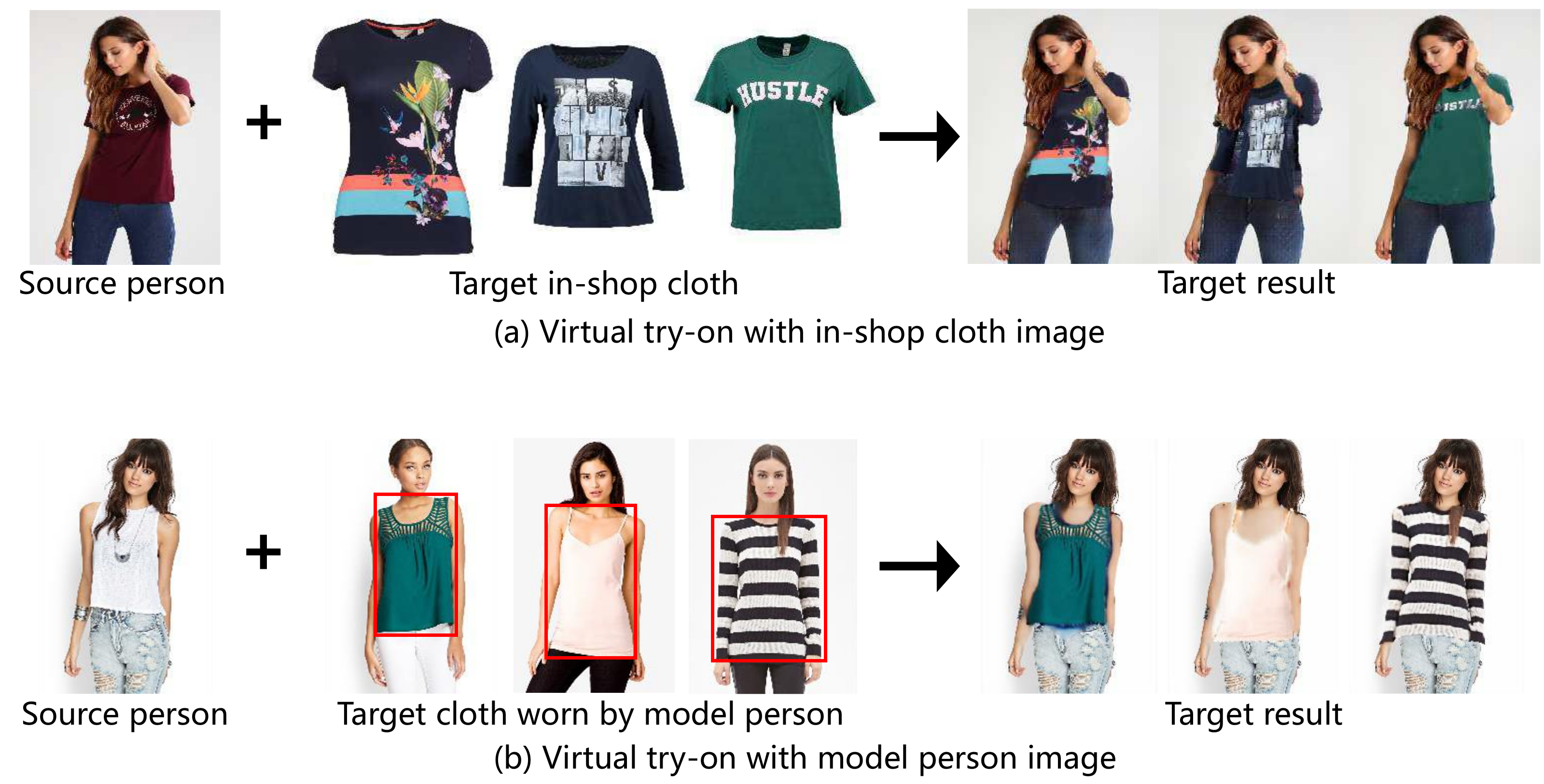}
	\caption{Illustration of virtual try-on. The target cloth is given as an in-shop image (a), or worn by a model person (b).}
	\label{HGT_overview}
\end{figure}

\begin{figure}[t]
	\centering
	\includegraphics[width=0.99\textwidth]{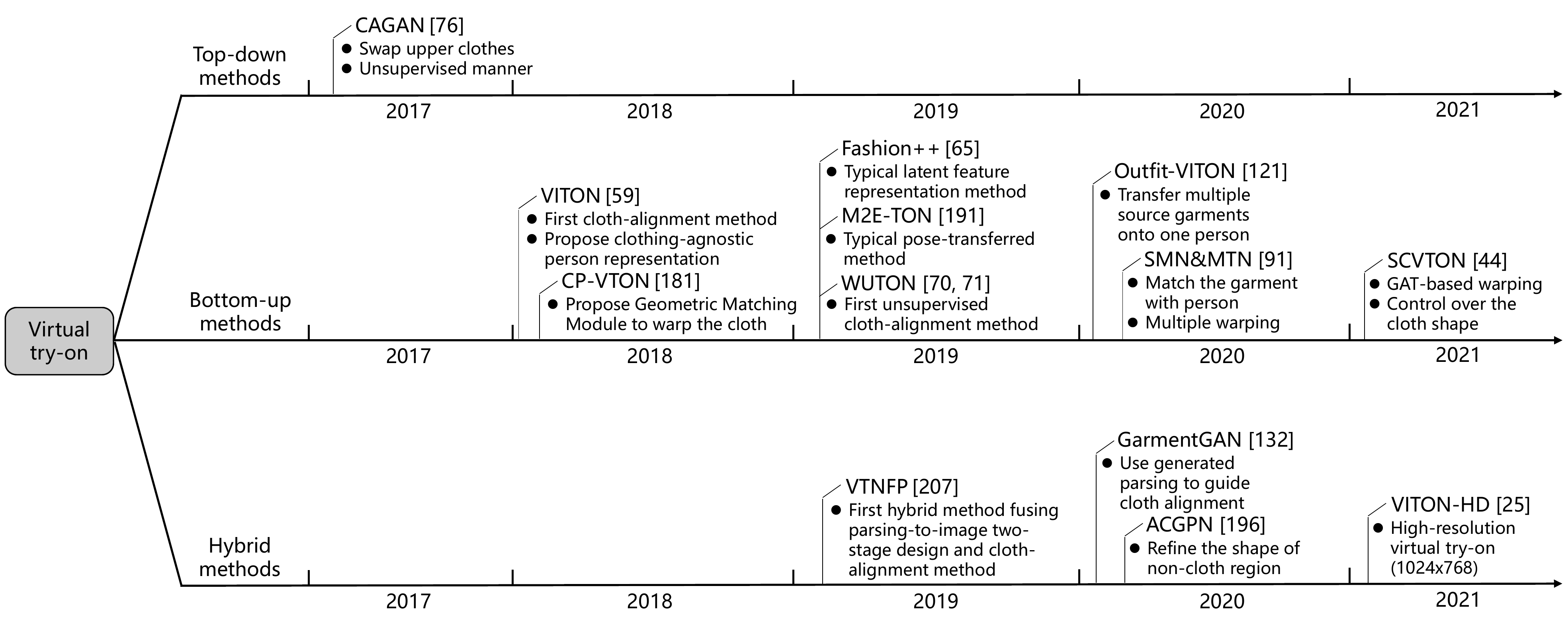}
	\caption{Representative works on virtual try-on.}
	\label{HGT representation}
\end{figure}

\begin{figure}[t]
	\centering
	\includegraphics[width=0.99\textwidth]{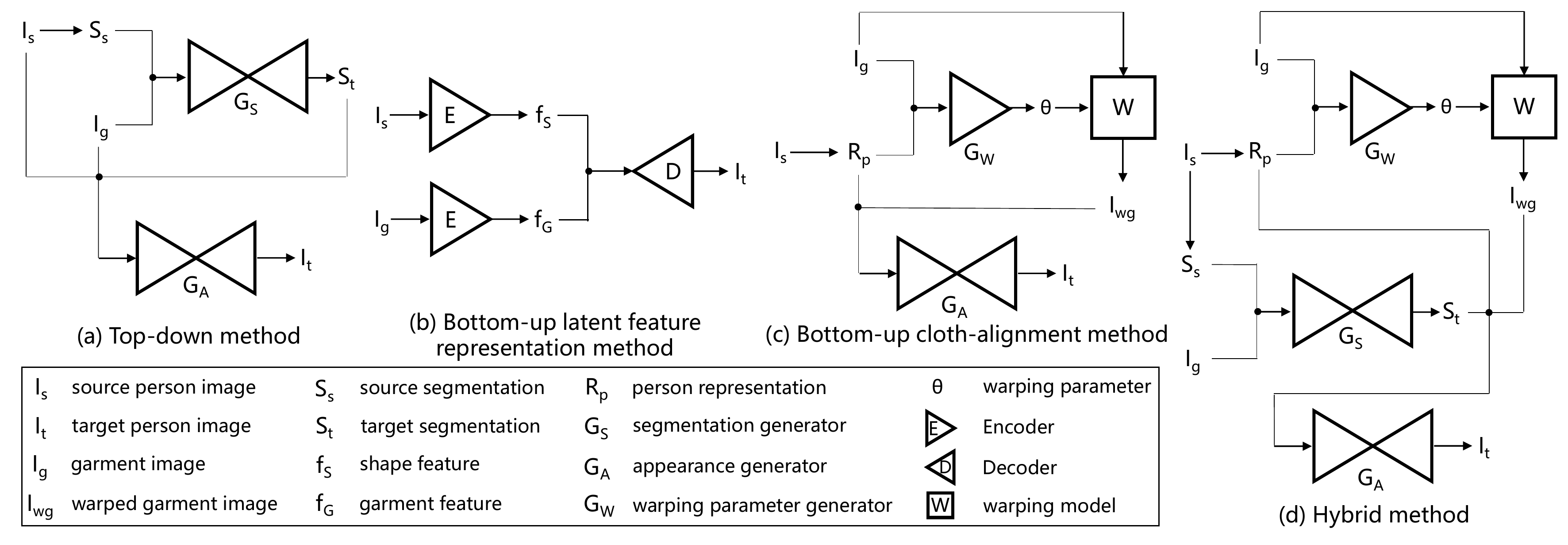}
	\caption{Comparison among four virtual try-on methods. (a) Top-down methods directly input $I_s$, $I_g$, $S_t$ to synthesize output $I_t$. (b) Bottom-up latent feature representation methods devote to represent the shape, garment and other features for synthesis. (c) Bottom-up cloth-alignment methods warp $I_g$ into the shape of $I_s$ via a warping model. (d) Hybrid methods combine top-down with the cloth-alignment method.}
	\label{HGT difference}
\end{figure}

\begin{table}[ht]
	\centering
	\caption{Summary of virtual try-on methods. Applicable garment: applicable types of input garment.}
	\scalebox{0.7}{
	\begin{tabular}{p{2.5cm}|p{8.7cm}|p{4.5cm}}
		\hline
		References & Main idea & Applicable garment \\
		\hline
		\hline
		\multicolumn{3}{c}{Top-down methods} \\
		\hline
		Raj \cite{SwapNet} & Parsing-to-image two-step generation & Garment of model person image \\
		\hline
		\cite{CAGAN, VTOwithClothingRegion} & Cycle consistency & In-shop upper cloth image \\
		\hline
		\multicolumn{3}{c}{Bottom-up latent feature representation methods} \\
		\hline
		Hsiao \cite{FashionPP} & Garment and pose feature representation and parsing-to-image two-step generation & Multiple garment images \\
		\hline
		Neuberger \cite{OutfitVITON} & Shape and garment feature representation parsing-to-image two-step generation & Multiple garment images \\
		\hline
		\multicolumn{3}{c}{Bottom-up cloth-alignment methods} \\
		\hline
		\cite{VITON, SPVITON} & Cloth-alignment coarse-to-fine generation & In-shop upper cloth image \\
		\hline
		\cite{CPVTON, CPVTONp} & Cloth alignment & In-shop upper cloth image \\
		\hline
		Lee \cite{LAVITON} & Cloth alignment and coarse-to-fine generation & In-shop upper cloth image \\
		\hline
		Kubo \cite{UVTON} & Coarse-to-fine 3D pose surface painting & In-shop upper cloth image \\
		\hline
		Han \cite{ClothFlow} & Segmentation-based clothing flow estimation & In-shop upper cloth image \\
		\hline
		Issenhuth \cite{TWUTON, SWUTON} & Cloth alignment and knowledge distillation & In-shop upper cloth image \\
		\hline
		Li \cite{SMVTON} & Garment-person pair matching and multi-warp & Upper cloth image \\
		\hline
		Xie \cite{ClothDecomposedVTON} & Decomposed cloth alignment & In-shop upper cloth image \\
		\hline
		Fincato \cite{VITONGT} & Two-stage cloth alignment & In-shop upper cloth image \\
		\hline
		Gao \cite{SCVTON2} & Graph Attention Network based cloth alignment & In-shop upper cloth image \\
		\hline
		Ge \cite{PFAFN} & Appearance flow based cloth alignment & In-shop upper cloth image \\
		\hline
		Ge \cite{DCTON} & Cycle consistency and cloth alignment & In-shop upper cloth image \\
		\hline
		Ren \cite{CIT} & Transformer based cloth alignment & In-shop upper cloth image \\
		\hline
		Liu \cite{SwapGAN} & Human pose transfer and cycle consistency & Garment of model person image \\
		\hline
		\cite{M2ETON, LiquidGAN} & Deformation method of human pose transfer & Garment of model person image \\
		\hline
		Yu \cite{IVTON} & Pose-transferred parsing-to-image generation & Garment of model person image \\
		\hline
		Roy \cite{LGVTON} & Pose transformation and coarse-to-fine generation & Garment of model person image \\
		\hline
		Cui \cite{DiOr} & Global flow field based cloth alignment & Garment of model person image \\
		\hline
		\multicolumn{3}{c}{Hybrid methods} \\
		\hline
		\cite{VTNFP, GarmentGAN} & Cloth-alignment parsing-to-image generation & In-shop upper cloth image \\
		\hline
		\cite{SieveNet, SCVTON, ACGPN, C3DVTON, CloTHVTON} & Cloth-alignment parsing-to-image generation and non-cloth region refinement & In-shop upper cloth image \\
		\hline
		Choi \cite{VITONHD} & High-resolution cloth-alignment parsing-to-image generation & In-shop upper cloth image \\
		\hline
	\end{tabular}}
	\label{HGT summary}
\end{table}

Virtual try-on, as shown in Fig.~\ref{HGT_overview}, is to transfer a specific garment onto a person image, which allows customers to virtually try on garments before online shopping. The early method \cite{VITONonWeb} focuses on 3D body and clothing reconstruction, mostly via manual modeling. This method is complex in procedure, and its results are less visually realistic. Since 2017, GAN-based approaches start to emerge as a complement to the 3D solution. As a comparison, 2D methods have much simple architecture and produce more realistic results. 

A brief summary of related works is shown in Tab.~\ref{HGT summary}. Deep-learning based virtual try-on can also be grouped into three categories: top-down, bottom-up and hybrid methods. Fig.~\ref{HGT representation} shows the representative methods, and Fig.~\ref{HGT difference} highlights the differences among the three branches.

\subsubsection{Top-down Methods}
\label{HGT Top-down Methods}

Top-down methods deal with virtual try-on from a global perspective. 
The straightforward way is to directly map the source person image to the final result, conditioning on the target garment. However, this approach only obtains a rough result. 
To better preserve shape and texture, human parsing maps are added as another conditioning input \cite{SwapNet, CAGAN, VTOwithClothingRegion}. 

In general, paired data is necessary for decent results, i.e., two images with the same identity and pose but with different garments. This assumption poses a big barrier in real scenarios. Therefore, unsupervised methods that operate on unpaired data, are more practical in real cases. 
Raj \textit{et al}. \cite{SwapNet} propose SwapNet that swaps garments between a pair of person images while preserving the pose and body shape. Self-supervised training is also applied to address the lack of paired data. 
Jetchev and Bergmann \cite{CAGAN} introduce a Conditional Analogy GAN (CAGAN) to transfer the upper clothes of a person into the target one, where cycle-consistency \cite{CycleGAN} is adopted in training. 
Kubo \textit{et al}. \cite{VTOwithClothingRegion} improve over CAGAN \cite{CAGAN} by segmenting the clothing region for better precision.

Similar to the pose-guided person generation, top-down virtual try-on methods are also difficult to preserve sufficient appearance details and garment texture details. Therefore, more people focus on bottom-up methods.

\subsubsection{Bottom-Up Methods}
\label{HGT Bottom-up Methods}

Top-down methods suffer from poor texture details. Thus two typical solutions are explored to improve the clothing details, namely latent feature representation and cloth-alignment.

\textbf{Latent Feature Representation Methods} \cite{ClothNet, FiNet, FashionPP, OutfitVITON} represent clothing attributes (e.g. shape and appearance) with latent features to better control the process of generation. These methods are widely adopted in tasks involving multiple garments.
Neuberger \textit{et al}. \cite{OutfitVITON} propose Outfit-VITON to try on multiple garments simultaneously, where the shape and appearance features are extracted to represent selected clothes. 
Hsiao \textit{et al}. \cite{FashionPP} propose Fashion++ to slightly adjust full-body clothing outfits, where multiple texture and shape features encode different parts of the garment and body.
Although the strong decomposition of latent feature representation methods is suitable for virtual try-on of multiple garments, some details are inevitably lost in the process of feature extraction.

\textbf{Cloth-alignment Methods} warp the target cloth into the shape of the source person. Due to its simplicity and high performance, cloth alignment makes the most popular bottom-up branch. According to the alignment strategies, cloth-alignment methods can be further divided into cloth-based warping and pose-based warping methods.

\textit{Cloth-based warping methods} directly deform cloth images with geometric transformation, which are generally applied to in-shop cloth images as shown in Fig.~\ref{HGT_overview}(a). 
The first method is introduced in VITON \cite{VITON}, which aligns clothes considering the pose, body shape and head region. 
Wang \textit{et al}. \cite{CPVTON} improve VITON by introducing Characteristic-Preserving Virtual Try-On Network (CP-VTON), which applies a Geometric Matching Module (GMM) to better align clothes. Subsequent methods are mostly following this paradigm.

Several works improve VITON and CP-VTON in different ways. 
Some of these methods focus on improving the accuracy of warping.
Lee \textit{et al}. \cite{LAVITON} and Fincato \textit{et al}. \cite{VITONGT} extend GMM to align clothes by a two-step warping. 
CP-VTON+ \cite{CPVTONp} explicitly regresses the warped cloth mask to improve the precision of alignment. 
Li \textit{et al}. \cite{SMVTON} propose a Shape-Matching-Net to choose shape-wise compatible ``garment-person'' pairs. 
Xie \textit{et al}. \cite{ClothDecomposedVTON} estimate the landmarks of the target cloth and warp the left sleeve, right sleeve and middle body, respectively.
Gao \textit{et al}. \cite{SCVTON2} transform the cloth image into the mesh and warp it with Graph Attention network (GAT) \cite{GAT}. This method gives control over the cloth shape, including length and tightness.
Some works introduce new techniques to optimize the results.
Issenhuth \textit{et al}. \cite{TWUTON, SWUTON} propose a Warping U-net for virtual Try-On Net (WUTON), which does not need the ground-truth target image during training. Meanwhile, they train a parsing-free student WUTON based on teacher-student knowledge distillation \cite{KnowledgeDistillation} strategy. 
Furthermore, Ge \textit{et al}. \cite{PFAFN} add a tutor model based on WUTON to train a more precise parsing-free virtual try-on model named PFAFN.
Ren \textit{et al}. \cite{CIT} introduce Transformer \cite{Transformer} to cloth alignment to capture long-range relation between garment and person.

Cloth-based warping methods can only deform clothing images based on body pose and shape, so it is difficult to handle the situations of complex body poses and occlusions.

\textit{Pose-based warping methods} take cloth images as a part of a person body, to deform based on body pose.
UVTON \cite{UVTON} deforms on the 3D dense pose \cite{DensePose} and then fills the surface with clothing textures. Han \textit{et al}. \cite{ClothFlow} propose to warp based on clothing flow.

This idea is applicable to the situation where reference cloth is worn by a model person as shown in Fig.~\ref{HGT_overview}(b). It greatly expands the application scope of cloth-alignment methods. Pose-guided generation (Sec.~\ref{ssec:pgpg}) is often adopted to register the model and source cloth.
Liu \textit{et al}. \cite{SwapGAN} introduce SwapGAN to directly transfer the model person to the target pose. Wu \textit{et al}. \cite{M2ETON} introduce an M2E Try-On Net (M2E-TON) to deform based on pose supervision. 
Roy \textit{et al}. \cite{LGVTON} introduce LGVTON to warp upper cloth using both pose and fashion landmarks \cite{FashionLandmark}. I-VTON \cite{IVTON} combines warped textures of multiple body parts for virtual try-on and then inpaints the missing parts with realistic appearances. 
Cui \textit{et al}. \cite{DiOr} estimate the global flow field (between source person and model person poses) to warp the garment, which can put on multiple garments in a certain order.

\subsubsection{Hybrid Methods}

Top-down methods can well capture the overall structure of a person and garments, but lose a lot of texture details. Bottom-up cloth-alignment methods easily preserve details on the garments, but the overall structure looks less natural. Therefore, some works seek a hybrid solution combining both perspectives.

Some methods \cite{VTNFP, GarmentGAN, VITONHD} take as input the human parsing map and aligned clothes, such that the top-down model is also aware of the internal structure used in bottom-up approaches. Moreover, Choi \textit{et al}. \cite{VITONHD} design an ALIAS generator to synthesize high-resolution images (1024$\times$768). Other methods \cite{SieveNet, SCVTON, ACGPN, C3DVTON, CloTHVTON} preserve details in the non-cloth region, also guided by the parsing map. Yang \textit{et al}. \cite{ACGPN} propose an Adaptive Content Generating and Preserving Network (ACGPN) that adds a non-target composition to refine the shape of the non-cloth region. Minar \textit{et al}. \cite{C3DVTON, CloTHVTON} align clothes based on a 3D clothing model. Meanwhile, they further consider non-cloth regions (e.g., skin, lower body) to comprehensively preserve details on the whole body.

\subsection{Garment Manipulation}
Besides virtual try-on, there are several tasks for garment manipulation, such as garment synthesis, text-guided garment manipulation and garment inpainting.

Garment synthesis generates person images with new garments. ClothNet \cite{ClothNet} generates photo-realistic clothing images from sketch with image-to-image translation \cite{pix2pix}. 
Text-guided garment manipulation aims to modify clothes with the guidance of text. Zhu \textit{et al}. \cite{FashionGAN} propose FashionGAN to generate clothing according to text descriptions. 
Garment inpainting uses in-painting methods to add garment textures. Han \textit{et al}. \cite{FiNet} present the FiNet to inpaint the missing garments such as upper/lower garments and shoes.

\subsection{Discussion}
\label{HGT discussion}


Virtual try-on is challenging for two main reasons. The first issue comes from the diversity of human poses, as well as the mismatch between garments and poses. Top-down methods can generate photo-realistic results for a wide range of poses, but can not well preserve the texture details. Latent feature representation methods support multi-garment virtual try-on, but are difficult to keep enough cloth textures. Cloth-alignment methods preserve most cloth textures, but can hardly handle the large mismatch between clothes and poses. In the future, we can combine latent feature representation with cloth alignment for predicting the clothing area as accurately as possible while retaining more details. The second issue is the lack of proper datasets for fully supervised training. Paired data is difficult to collect in the scenario of virtual try-on. But, the obstacle in supervised data gives rise to unsupervised methods such as \cite{CAGAN, TWUTON, OutfitVITON}. 

So far, bottom-up methods are more popular. Due to the advantage of retaining texture details, cloth-alignment methods are developing rapidly. To improve the generation quality, many deformation methods have been explored, including decomposed TPS \cite{ClothDecomposedVTON} and ClothFlow \cite{ClothFlow}. 
Besides, the variety of garments for try-on is getting much wider recently. Early works only support upper clothes. Recent works \cite{FashionPP, OutfitVITON} begin to try on clothes including upper clothes, trousers, skirts, hats, etc. However, current methods can not work well on wild datasets and multi-garments.

\section{Benchmarks}

\label{Datasets and Metrics}
We first review the datasets and evaluation metrics in this section, and then benchmark recent performances where applicable. 

\subsection{Datasets}

\begin{table}[ht]
	\centering
	\caption{Summary of talking-head video datasets.}
	\scalebox{0.65}{
	\begin{tabular}{p{2.3cm}|p{1.8cm}|c|c|c|c|c|c}
		\hline
		Name & Year & Data scale (in hours) & \# Speaker & \# Sentence & Head movement & Emotion & Open source \\
		\hline
		\hline
		GRID & 2006 \cite{GRID} & 27.5 & 33 & 33k & $\times$ & $\times$ & \href{https://www.grid.ac/}{Link} \\
		\hline
		TCD-TIMIT & 2015 \cite{TCDTIMIT} & 11.1 & 62 & 6.9k & $\times$ & $\times$ & \href{https://sigmedia.github.io/resources/dataset/tcd_timit/}{Link} \\
		\hline
		LRW & 2016 \cite{LRW} & 173 & 1k+ & 539k & $\times$ & $\times$ & \href{https://www.robots.ox.ac.uk/~vgg/data/lip_reading/lrw1.html}{Link} \\
		\hline
		MODALITY & 2017 \cite{MODALITY} & 31 & 35 & 5.8k & $\times$ & $\times$ & \href{http://www.modality-corpus.org/}{Link} \\
		\hline
		CREMA-D & 2014 \cite{CREMA-D} & 11.1 & 91 & 12 & $\sqrt{}$ & $\sqrt{}$ & \href{https://github.com/CheyneyComputerScience/CREMA-D}{Link} \\
		\hline
		MSP-IMPROV & 2016 \cite{MSPIMPROV} & 18 & 12 & 652 & $\sqrt{}$ & $\sqrt{}$ & \href{https://ecs.utdallas.edu/research/researchlabs/msp-lab/MSP-Improv.html}{Link} \\
		\hline
		ObamaSet & 2017 \cite{SynthesisObama} & 14 & 1 & --- & $\sqrt{}$ & $\times$ & \href{https://www.youtube.com}{Link} \\
		\hline
		VoxCeleb & 2017 \cite{VoxCeleb} & 352 & 1.2k & 153.5k & $\sqrt{}$ & $\times$ & \href{https://www.robots.ox.ac.uk/~vgg/data/voxceleb/}{Link} \\
		\hline
		VoxCeleb2 & 2018 \cite{VoxCeleb2} & 2.4k & 6.1k & 1.1m & $\sqrt{}$ & $\times$ & \href{https://www.robots.ox.ac.uk/~vgg/data/voxceleb/vox2.html}{Link} \\
		\hline
		RAVDESS & 2018 \cite{RAVDESS} & 7 & 24 & 2 & $\sqrt{}$ & $\sqrt{}$ & \href{https://www.kaggle.com/datasets/uwrfkaggler/ravdess-emotional-speech-audio}{Link} \\
		\hline
		LRS2-BBC & 2018 \cite{LRS2BBC} & 224.5 & 500+ & 140k+ & $\sqrt{}$ & $\times$ & \href{https://www.robots.ox.ac.uk/~vgg/data/lip_reading/lrs2.html}{Link} \\
		\hline
		LRS3-TED & 2018 \cite{LRS3TED} & 438 & 5k+ & 152k+ & $\sqrt{}$ & $\times$ & \href{https://www.robots.ox.ac.uk/~vgg/data/lip_reading/lrs3.html}{Link} \\
		\hline
		MELD & 2018 \cite{MELD} & 13.7 & 407 & 13.7k & $\sqrt{}$ & $\sqrt{}$ & \href{https://affective-meld.github.io/}{Link} \\
		\hline
		Lombard & 2018 \cite{Lombard} & 3.6 & 54 & 5.4k & $\sqrt{}$ & $\sqrt{}$ & \href{https://spandh.dcs.shef.ac.uk//avlombard/}{Link} \\
		\hline
		Faceforensics++ & 2019 \cite{Faceforensicspp} & 5.7 & 1k & 1k+ & $\sqrt{}$ & $\times$ & \href{https://github.com/ondyari/FaceForensics}{Link} \\
		\hline
		MEAD & 2020 \cite{MEAD} & 39 & 60 & 20 & $\sqrt{}$ & $\sqrt{}$ & \href{https://github.com/uniBruce/Mead}{Link} \\
		\hline
	\end{tabular}}
	\label{Talking head datasets}
\end{table}

\begin{table}[ht]
	\centering
	\caption{Datasets for pose and garment-oriented person generation. PGPIG: Pose-Guided Person Image Generation. PGPVG: Pose-Guided Person Video Generation. VTON: Virtual Try-on. VF: Virtual Fitting. VVTON: Video Virtual Try-on.}
	\scalebox{0.65}{
	\begin{tabular}{p{2cm}|p{1.8cm}|p{7.5cm}|p{3.3cm}|c}
		\hline
		Name & Year & Data scale & Applicable fields & Open source \\
		\hline
		\hline
		Market-1501 & 2015 \cite{Market1501} & 32,668 images of 1,501 persons & PGPIG & \href{https://zheng-lab.cecs.anu.edu.au/Project/project_reid.html}{Link} \\
		\hline
		DeepFasion & 2016 \cite{DeepFashion} & 52,712 in-shop cloth images and over 200,000 cross-pose / scale pairs & PGPIG and VTON & \href{https://mmlab.ie.cuhk.edu.hk/projects/DeepFashion.html}{Link} \\
		\hline
		MVC & 2016 \cite{MVC} & 161,638 clothing images of 37,499 items & PGPIG and VTON & \href{https://mvc-datasets.github.io/MVC/}{Link} \\
		\hline
		Human3.6M & 2013 \cite{Human36M} & 3,578,080 images of 11 persons & PGPIG & \href{http://vision.imar.ro/human3.6m/description.php}{Link} \\
		\hline
		Chictopia10k & 2015 \cite{Chictopia10k} & 17,706 images & VTON & \href{https://files.is.tuebingen.mpg.de/classner/gp/}{Link} \\
		\hline
		Zalando & 2018 \cite{VITON} & 16,253 person-cloth pairs & VTON & NA \\
		\hline
		LookBook & 2016 \cite{LookBook} & 75,016 person images and 9,732 in-shop cloth images & PGPIG and VTON & NA \\
		\hline
		MPV & 2019 \cite{MGVTON} & 35,687 person images and 13,524 in-shop cloth images & PGPIG, VTON and VF & \href{https://competitions.codalab.org/competitions/23471}{Link} \\
		\hline
		FashionOn & 2019 \cite{FashionOn} & 21,790 person images and 10,895 in-shop cloth images & PGPIG, VTON and VF & NA \\
		\hline
		FashionTryOn & 2019 \cite{PGVTON} & 57,428 person images and 28,714 in-shop cloth images & PGPIG, VTON and VF & \href{https://fashiontryon.wixsite.com/fashiontryon}{Link} \\
		\hline
		Penn Action & 2013 \cite{PennAction} & 2,326 videos & PGPIG and PGPVG & \href{http://dreamdragon.github.io/PennAction/}{Link} \\
		\hline
		Tai-Chi & 2018 \cite{MoCoGAN} & 4,500 videos & PGPIG and PGPVG & \href{https://github.com/AliaksandrSiarohin/first-order-model/blob/master/data/taichi-loading/README.md}{Link} \\
		\hline
		Fashion & 2019 \cite{DwNet} & 600 videos & PGPIG and PGPVG & \href{https://vision.cs.ubc.ca/datasets/fashion/}{Link} \\
		\hline
		iPER & 2019 \cite{LiquidGAN} & 206 videos of 30 persons & PGPIG, VTON and PGPVG & \href{https://svip-lab.github.io/dataset/iPER_dataset.html}{Link} \\
		\hline
		VVT & 2019 \cite{FWGAN} & 791 videos, 791 person images and 791 cloth images & PGPIG, VTON, PGPVG and VVTON & \href{https://competitions.codalab.org/competitions/23472}{Link} \\
		\hline
	\end{tabular}}
	\label{HPT HGT datasets}
\end{table}

Details of popular datasets for talking-head generation are summarized in Tab.~\ref{Talking head datasets}, and those for pose and garment-oriented generation are in Tab.~\ref{HPT HGT datasets}. Here we only briefly highlight the datasets adopted by the major researchers.

For talking-head generation, GRID \cite{GRID} and LRW \cite{LRW} are adopted to evaluate methods without head motion; CREMA-D \cite{CREMA-D} validates on videos with spontaneous motions; VoxCeleb2 \cite{VoxCeleb2} and LRS3-TED \cite{LRS3TED} focus on in-the-wild videos. Alternatively, Chen \textit{et al.} \cite{StandardEvaluation} also suggest several evaluation protocols for evaluating talking heads generation.

For pose-guided person image generation, the most commonly adopted datasets are Market-1501 \cite{Market1501} and DeepFasion \cite{DeepFashion}. For garment-based generation, Zalando \cite{VITON} and DeepFasion \cite{DeepFashion} datasets are popular for evaluation. Recently, MPV \cite{MGVTON} dataset is also used to evaluate garment and pose-based generation.

\subsection{Evaluation Metrics}
Evaluating generation tasks are known difficult. As a common practice, multiple objectives (e.g, Inception Score, SSIM) and subjectives (e.g., Amazon Mechanical Turk, User study) metrics are adopted for a comprehensive evaluation. 
Subjective evaluation involves humans in the loop, which is often applied to compare the perceptual visual quality of generated content. However, due to the subjective factors and higher costs during evaluation, most works also seek quantitative evaluations with objective metrics. For person generation, the common objective metrics are summarized below.

\textbf{SSIM} (Structural Similarity) \cite{SSIM} measures the quality of generated image compared with the original image, which is widely used in image synthesis. Specifically, SSIM calculates the similarity between the synthesized image and the ground-truth real image in three dimensions luminance, contrast and structure. 

\textbf{IS} (Inception Score) \cite{IS} evaluates the quality of generated image in terms of clear objects and high diversity. IS calculates the distribution of generated images via a pre-trained Inception v3 \cite{InceptionV3} network. This metric is not that informative, and Barratt and Sharma \cite{NoteOfIS} present some shortcomings of IS.

\textbf{FID} (Fr\'echet Inception Distance) \cite{FID} calculates the Fr\'echet distance between the distribution of generated images and real images to capture their similarity.

\textbf{FReID} uses a pre-trained person re-ID model to estimate the Gaussian distribution of generated images and real images, which calculates the Fr\'echet distance of these distributions.

\textbf{LPIPS} (Learned Perceptual Image Patch Similarity) \cite{LPIPS} uses a pre-trained deep network to learn deep perceptual features of the image patch and then computes the average $\ell_2$ distance between features of two images. 

\textbf{DS} (Detection Score) measures the confidence of a pre-trained person detector in a generated person image, by computing the average person-class detection scores on generated images.

\textbf{AttrRec-k} (Top-k Clothing Attribute Retaining Rate) uses a pre-trained clothing attribute recognition model to predict clothing attributes from a generated image. It utilizes the top-k recall rate as the final score.

\subsection{Performance Comparison}
Talking-head video generation is mostly evaluated by user studies.
Recently, Chen \textit{et al.} \cite{StandardEvaluation} propose several objective metrics to evaluate talking-head videos in terms of identity preserving, lip synchronization, video quality, and spontaneous head movements. Meanwhile, they present a new benchmark with standardized dataset pre-processing strategies for evaluating talking-head generation methods.

\begin{table}[b]
	\centering
	\caption{Comparison of state-of-the-art methods for pose-guided person image generation. $\uparrow$: larger is better. }
	\scalebox{0.65}{
	\begin{tabular}{p{2.3cm}|c|c|c|c|c|c|c|c|c}
		\hline
		\multirow{2}*{Methods} & \multirow{2}*{Year} & \multicolumn{5}{c}{Market-1501} & \multicolumn{3}{|c}{DeepFashion} \\
		\cline{3-10}
		~ & ~ & SSIM$\uparrow$ & IS$\uparrow$ & mask-SSIM$\uparrow$ & mask-IS$\uparrow$ & FID$\downarrow$ & SSIM$\uparrow$ & IS$\uparrow$ & FID$\downarrow$ \\
		\hline
		Zhao \cite{VariGAN} & 2017 & --- & --- & --- & --- & --- & 0.62 & 3.03 \cite{DPT} & --- \\
		\hline
		Ma \cite{PG2} & 2017 & 0.253 & 3.460 & 0.792 & 3.435 & --- & 0.762 & 3.090 & --- \\
		\hline
		Pumarola \cite{UPIS} & 2018 & --- & --- & --- & --- & --- & 0.747 & 2.97 & --- \\
		\hline
		Esser \cite{VUnet} & 2018 & 0.353 & 3.214 & --- & --- & 20.144 \cite{DIST} & 0.786 & 3.087 & 23.667 \cite{DIST} \\
		\hline
		Ma \cite{Disentangle} & 2018 & 0.099 & 3.483 & 0.614 & 3.491 & --- & 0.614 & 3.228 & --- \\
		\hline
		Siarohin \cite{DefGAN} & 2018 & 0.290 & 3.185 & 0.805 & 3.502 & 25.364 \cite{DIST} & 0.756 & 3.439 & 18.457 \cite{DIST} \\
		\hline
		Dong \cite{WarpingGAN} & 2018 & 0.356 & 3.409 & --- & --- & --- & 0.793 & 3.314 & --- \\
		\hline
		Neverova \cite{DPT} & 2018 & --- & --- & --- & --- & --- & 0.785 & 3.61 & --- \\
		\hline
		Sun \cite{Sun2019} & 2019 & 0.344 & 3.291 & --- & --- & --- & 0.789 & 3.006 & --- \\
		\hline
		Tang \cite{C2GAN} & 2019 & 0.282 & 3.349 & 0.811 & 3.510 & --- & --- & --- & --- \\
		\hline
		Song \cite{UPIGSPT} & 2019 & 0.203 & 3.499 & 0.758 & 3.680 & --- & 0.736 & 3.441 & --- \\
		\hline
		Siarohin \cite{AttentionMHIG} & 2019 & 0.326 & 3.613 & 0.806 & 3.814 & --- & 0.774 & 3.421 & --- \\
		\hline
		Liang \cite{PCGAN} & 2019 & --- & 3.657 & --- & 3.614 & 20.355 & --- & 3.536 & 29.684 \\
		\hline
		Han \cite{ClothFlow} & 2019 & --- & --- & --- & --- & --- & 0.771 & 3.88 & --- \\
		\hline
		Li \cite{DIAF} & 2019 & --- & --- & --- & --- & 27.163 \cite{DIST} & 0.778 & 3.338 & 16.314 \cite{DIST} \\
		\hline
		Dong \cite{PPGAN} & 2019 & 0.396 & 3.581 & --- & --- & --- & 0.782 & 3.595 & --- \\
		\hline
		Zhu \cite{PATN} & 2019 & 0.311 & 3.323 & 0.811 & 3.773 & 22.657 \cite{DIST} & 0.773 & 3.209 & 20.739 \cite{DIST} \\
		\hline
		Karmakar \cite{ptGAN} & 2020 & 0.302 & 3.488 & --- & --- & --- & 0.781 & 3.238 & --- \\
		\hline
		Ren \cite{DIST} & 2020 & --- & --- & --- & --- & 19.751 & --- & --- & 10.573 \\
		\hline
		Yang \cite{DRN} & 2020 & --- & --- & --- & --- & --- & 0.774 & 3.125 & 14.611 \\
		\hline
		Li \cite{PoNA} & 2020 & 0.315 & 3.487 & 0.814 & 3.867 & --- & 0.775 & 3.338 & --- \\
		\hline
		Chen \cite{PMAN} & 2021 & 0.306 & 3.827 & 0.804 & 3.809 & --- & 0.760 & 3.348 & --- \\
		\hline
		Liu \cite{SPAN} & 2021 & 0.732 & 3.703 & 0.821 & 3.750 & 16.142 & 0.786 & 3.736 & 8.732 \\
		\hline
	\end{tabular}}
	\label{HPT quantitative results}
\end{table}

\begin{table}[t]
	\centering
	\caption{Comparison of state-of-the-art virtual try-on methods. $\uparrow$: larger is better.}
	\scalebox{0.65}{
	\begin{tabular}{p{1.7cm}|c|c|c|c|c|c|c}
		\hline
		\multirow{2}*{Methods} & \multirow{2}*{Year} & \multicolumn{3}{c}{Zalando} & \multicolumn{3}{|c}{DeepFashion} \\
		\cline{3-8}
		~ & ~ & SSIM$\uparrow$ & IS$\uparrow$ & FID$\downarrow$ & SSIM$\uparrow$ & IS$\uparrow$ & FID$\downarrow$ \\
		\hline
		Han \cite{VITON} & 2018 & 0.786 \cite{ClothFlow} & 2.514 & 41.80 \cite{IVTON} & 0.71 \cite{LGVTON} & 2.40 \cite{SwapGAN} & 78.45 \cite{LGVTON} \\
		\hline
		Raj \cite{SwapNet} & 2018 & 0.83 & 2.631 \cite{IVTON} & 114.50 \cite{IVTON} & --- & --- & --- \\
		\hline
		Wang \cite{CPVTON} & 2018 & 0.792 \cite{ClothFlow} & 2.748 \cite{SPVITON} & 23.60 \cite{IVTON} & 0.72 \cite{LGVTON} & 2.41 \cite{LGVTON} & 72.95 \cite{LGVTON} \\
		\hline
		Song \cite{SPVITON} & 2019 & --- & 2.656 & --- & --- & --- & --- \\
		\hline
		Han \cite{ClothFlow} & 2019 & 0.803 & --- & --- & --- & --- & --- \\
		\hline
		Yu \cite{VTNFP} & 2019 & 0.803 \cite{ACGPN} & 2.784 \cite{ACGPN} & --- & --- & --- & --- \\
		\hline
		Liu \cite{SwapGAN} & 2019 & --- & --- & --- & 0.717 & 2.65 & --- \\
		\hline
		Wu \cite{M2ETON} & 2019 & --- & 2.510 \cite{IVTON} & 33.28 \cite{IVTON} & --- & --- & --- \\
		\hline
		Yu \cite{IVTON} & 2019 & --- & 2.708 & 29.48 & --- & --- & --- \\
		\hline
		Roy \cite{LGVTON} & 2020 & --- & --- & --- & 0.86 & 2.71 & 56.85 \\
		\hline
		Jandial \cite{SieveNet} & 2020 & 0.766 & 2.82 & 14.65 & --- & --- & --- \\
		\hline
		Raffie \cite{GarmentGAN} & 2020 & --- & 2.774 & 16.578 & --- & --- & --- \\
		\hline
		Yang \cite{ACGPN} & 2020 & 0.845 & 2.829 & 15.67 \cite{PFAFN} & --- & --- & --- \\
		\hline
		Minar \cite{CPVTONp} & 2020 & 0.817 & 3.074 & --- & --- & --- & --- \\
		\hline
		Fincato \cite{VITONGT} & 2021 & 0.886 & 2.76 & 12.45 & --- & --- & --- \\
		\hline
		Ge \cite{PFAFN} & 2021 & --- & --- & 10.09 & --- & --- & --- \\
		\hline
		Ge \cite{DCTON} & 2021 & 0.83 & 2.85 & 14.82 & --- & --- & --- \\
		\hline
		Ren \cite{CIT} & 2021 & 0.827 & 3.060 & --- & --- & --- & --- \\
		\hline
	\end{tabular}}
	\label{HGT quantitative results}
\end{table}

\begin{figure}[t]
	\centering
	\includegraphics[width=0.8\textwidth]{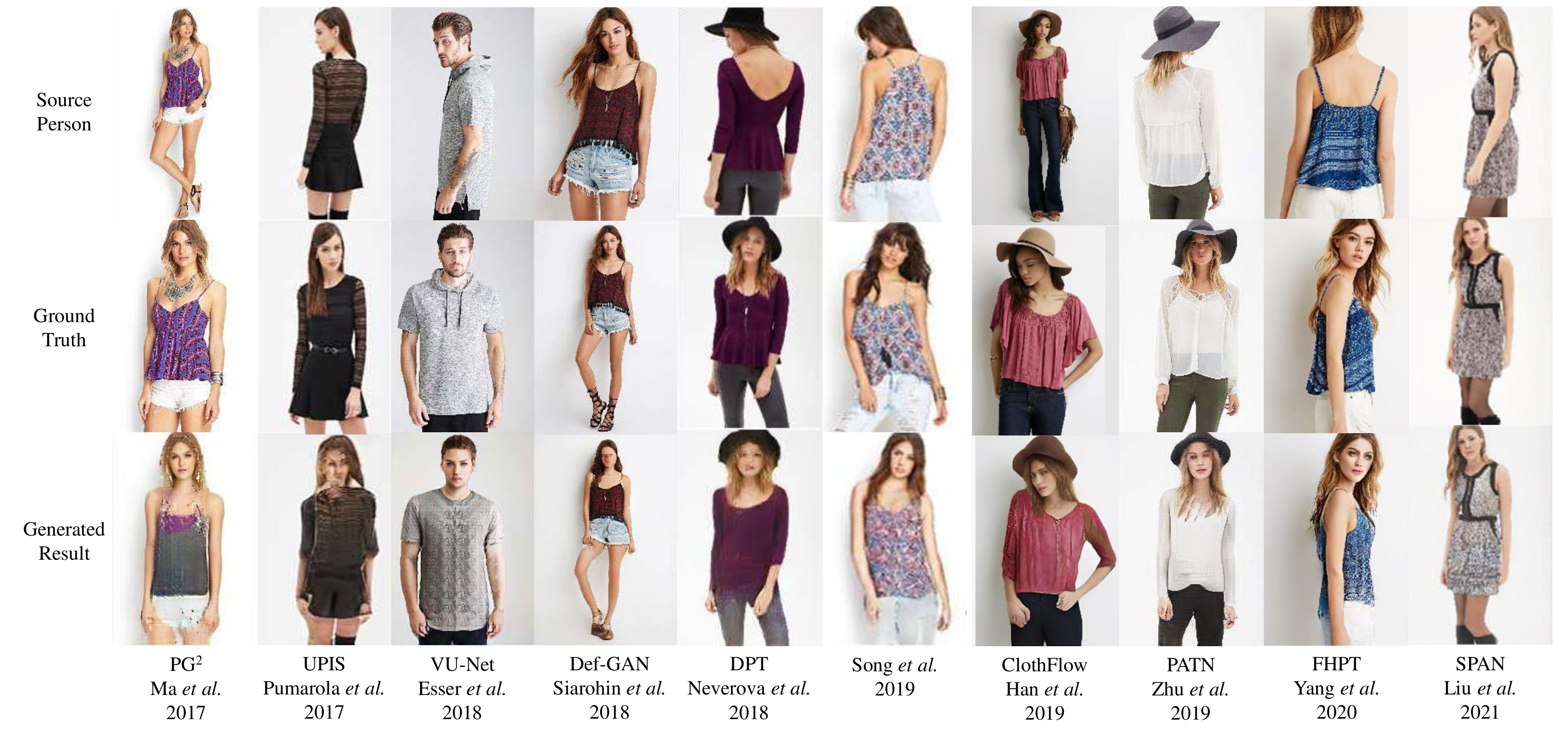}
	\caption{The qualitative comparisons with state-of-the-art pose-guided person image generation methods on DeepFashion \cite{DeepFashion} dataset, including PG$^{2}$ \cite{PG2}, UPIS \cite{UPIS}, VU-Net \cite{VUnet}, Def-GAN \cite{DefGAN}, DPT \cite{DPT}, Song \textit{et al.} \cite{UPIGSPT}, ClothFlow \cite{ClothFlow}, PATN \cite{PATN}, FHPT \cite{DRN} and SPAN \cite{SPAN}.}
	\label{Pose compare}
\end{figure}

\begin{figure}[t]
	\centering
	\includegraphics[width=0.9\textwidth]{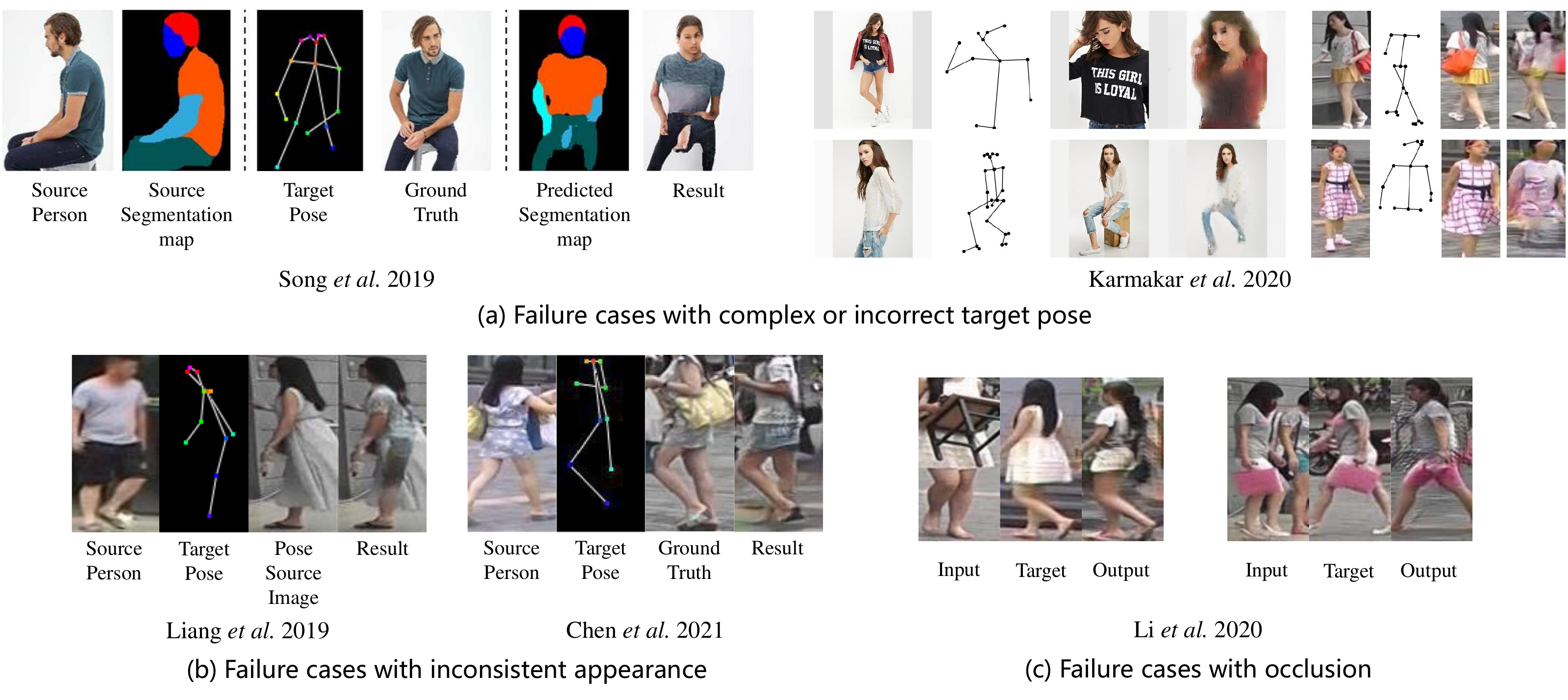}
	\caption{Some failure cases with (a) complex or incorrect target pose, (b) inconsistent appearance and (c) occlusion. These cases are from Song \textit{et al.} \cite{UPIGSPT}, Karmakar \textit{et al.} \cite{ptGAN}, Liang \textit{et al.} \cite{PCGAN}, Chen \textit{et al.} \cite{PMAN} and Li \textit{et al.} \cite{PoNA}.}
	\label{Pose fail}
\end{figure}

\begin{figure}[t]
	\centering
	\includegraphics[width=0.8\textwidth]{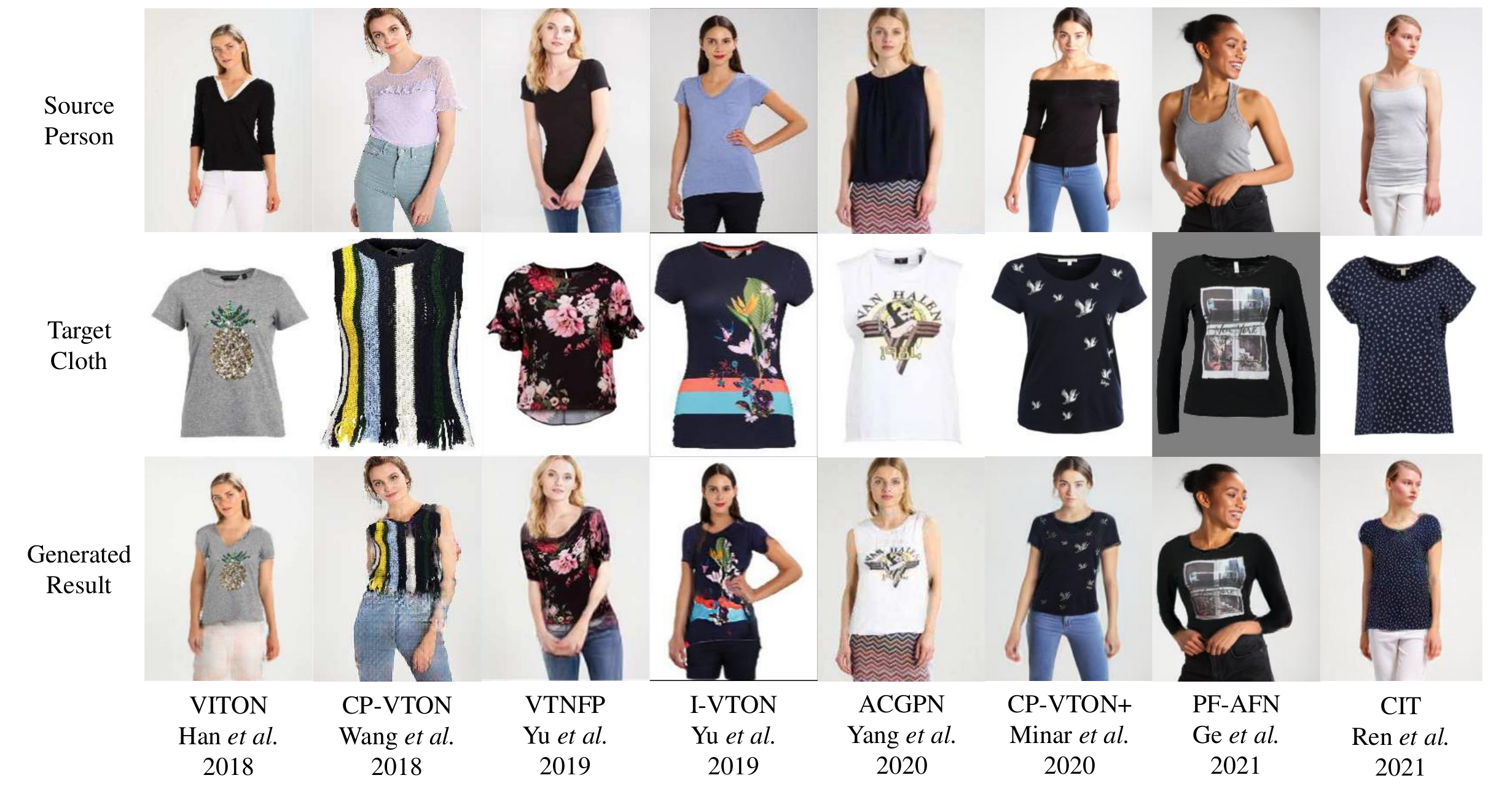}
	\caption{The qualitative comparisons with state-of-the-art virtual try-on methods on Zalando \cite{VITON} dataset, including VITON \cite{VITON}, CP-VTON \cite{CPVTON}, VTNFP \cite{VTNFP}, I-VTON \cite{IVTON}, ACGPN \cite{ACGPN}, CP-VTON+ \cite{CPVTONp}, PF-AFN \cite{PFAFN} and CIT \cite{CIT}.}
	\label{Garment compare}
\end{figure}

\begin{figure}[t]
	\centering
	\includegraphics[width=0.9\textwidth]{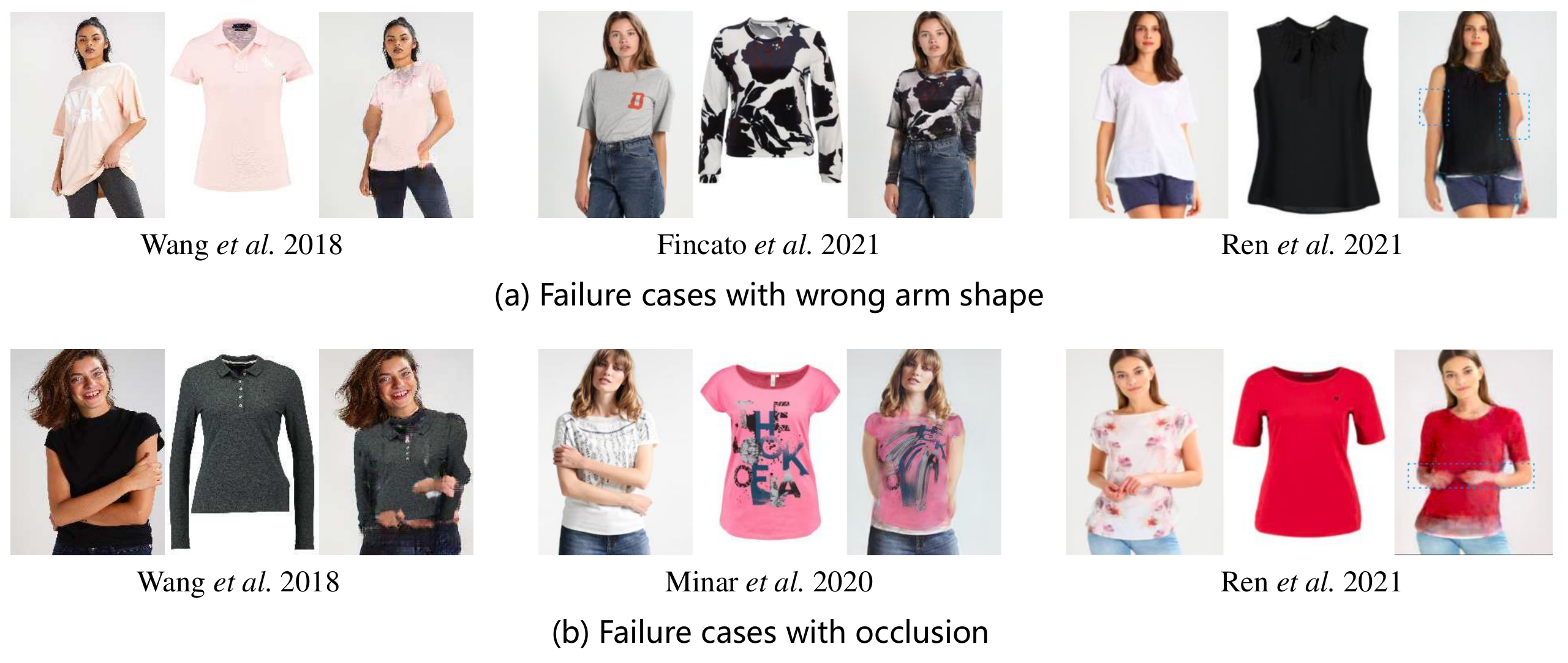}
	\caption{Some failure cases with (a) wrong arm shape and (b) occlusion. These cases are from Wang \textit{et al.} \cite{CPVTON}, Fincato \textit{et al.} \cite{VITONGT}, Minar \textit{et al.} \cite{CPVTONp} and Ren \textit{et al.} \cite{CIT}.}
	\label{Garment fail}
\end{figure}

For pose-guided person generation and virtual try-on, SSIM \cite{SSIM}, IS \cite{IS} and their variants are widely applied. But these metrics are not perfect \cite{LPIPS, NoteOfIS}. People gradually turn to FID \cite{FID} and LPIPS \cite{LPIPS} for evaluation. In particular, user studies are more widely used in virtual try-on.

Tab.~\ref{HPT quantitative results} shows the quantitative results of major state-of-the-art pose-guided person image generation methods. On average, deformation and hybrid methods are better than top-down and latent feature representation methods. For a more intuitive comparison, Fig.~\ref{Pose compare} shows the qualitative comparisons with some state-of-the-art methods. Meanwhile, Fig.~\ref{Pose fail} shows some failure cases. It demonstrates that the present pose-guided person image generation methods are still difficult to deal with complex poses and in-the-wild images.

Similarly, Tab.~\ref{HGT quantitative results} gives the comparison for virtual try-on methods. Hybrid methods are more competitive than other methods. Fig.~\ref{Garment compare} shows the visual comparison with some state-of-the-art methods. Meanwhile, Fig.~\ref{Garment fail} shows some failure cases. It indicates that the present virtual try-on methods are still difficult to handle different garment shapes and occlusion.

\section{Discussion} \label{Common Discussion}

\begin{figure}[t]
	\centering
	\includegraphics[width=0.7\textwidth]{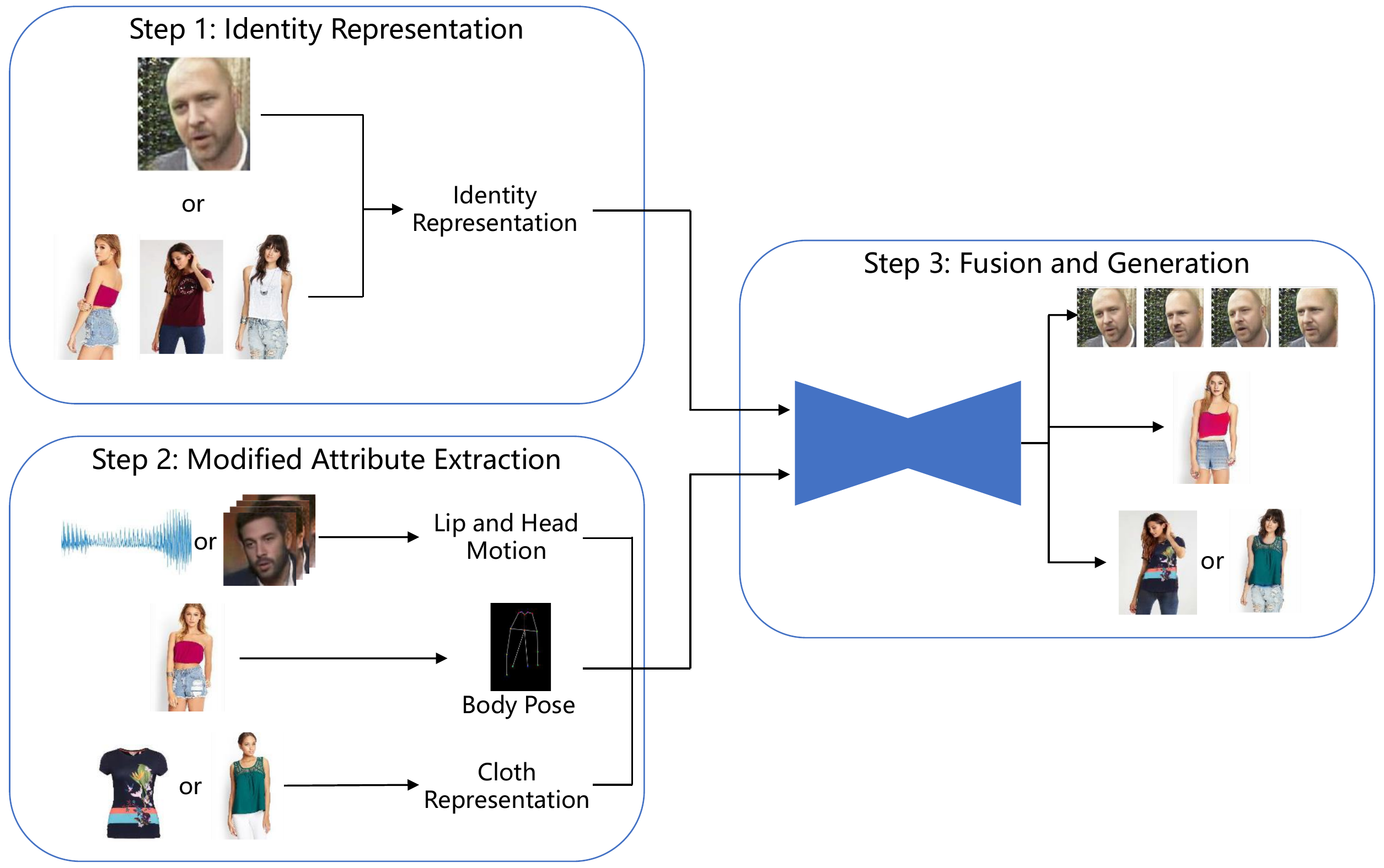}
	\caption{The framework of talking-head generation, pose-guided person generation and virtual try-on.}
	\label{common framework}
\end{figure}


In Sections 2, 3 and 4, we have reviewed talking-head generation, pose-guided person generation and virtual try-on, respectively. In this section, we will discuss the commonalities between these three tasks.

\begin{itemize}
\item \textbf{Same high-level framework.}
The three tasks all firstly construct a representation of the source person identity, extract and encode the modified attributes, and finally combine the identity representation and attribute encoding to obtain the target result. Fig.~\ref{common framework} illustrates this common framework. 

\item \textbf{The idea of deformation.}
There are many talking-head generation methods \cite{Portrait2Life, WarpGuidedGAN, FLNet, FaRGAN, FSGAN, ReenactGAN, MakeltTalk} that deform the source face image based on the target facial landmarks. Flow warping is the major deformation method. 

Deformation methods in pose-guided person generation deform the source person according to the target pose. The deformation tools include decoupled deformation, global flow warping and deformation based on the attention mechanism.

Cloth-alignment methods in virtual try-on also deform the target cloth according to the source person pose. TPS and flow warping are two major deformation tools.

These three types of methods are essentially the same: they all deform the image appearance according to the pose. As evidence, some methods \cite{C2GAN, DIST} use the same model for talking-head generation and pose-guided person generation. Meanwhile, some works \cite{LRW, ClothFlow} use the same methods for pose-guided person generation and virtual try-on simultaneously.

\item \textbf{The idea of feature representation and fusion.}
Since the three tasks all need to express the person identity and modified attribute, latent feature representation methods appear in all three tasks. To preserve the identity information, latent feature representation methods focus on feature decoupling to extract pure identity information and pure attribute information as far as possible.

Talking-head generation pays more attention to the decomposition degree, while pose-guided person generation and virtual try-on pay less attention. Talking-head generation needs to generate rich facial details. In order to preserve more identity information, interference from other attributes should be eliminated as much as possible. Meanwhile, pose-guided person generation and virtual try-on do not need to generate rich details related to identity information. However, with the development of research and the improvement of demand, it is inevitable to study how to improve the decomposition degree in pose-guided person generation and virtual try-on.

\end{itemize}
\section{Applications}
\label{Applications}
We have reviewed deep person generation from the perspective of face, pose and cloth synthesis. 
This section presents typical applications based on the aforementioned fundamental tasks. We mainly focus on three typical applications: Generative Data Augmentation (how generated data can help machines), Virtual Fitting and Digital Human (how generated data can help human beings).

\subsection{Generative Data Augmentation}
\label{Augmentation}


Most deep learning models are data-hungry for superior performance. 
Deep person generation is often adopted as data augmentation in person-related tasks such as person ReID (Re-identification), pedestrian detection and autonomous driving.

Person ReID requires massive images of the same person from different views, but existing datasets can only provide limited images for each ID. 
Zheng \textit{et al}. \cite{LSRO} generate person images with different views and propose a semi-supervised learning method to utilize the generated unlabeled data. Liu \textit{et al}. \cite{PTReID} use pose-guided generation to augment the ReID. Zhang \textit{et al}. \cite{MVPA} use an improved PG$^{2}$ \cite{PG2} network to synthesize person images with diverse poses, where the manual labeling is omitted.
There are also methods integrating the generation process into the ReID pipeline \cite{PNGAN, FDGAN, DGNet}.
Ge \textit{et al}. \cite{FDGAN} propose a Feature Distilling GAN (FD-GAN), where pose-guided generation is used to learn the robust identity-related and pose-unrelated features. Zheng \textit{et al}. \cite{DGNet} present a DG-Net to tactfully combine the re-ID discriminative module with a pose-guided generation module, which generates new person images in new poses and discriminates the ID of the generated images simultaneously.

For tasks in pedestrian detection and autonomous driving, person generation is also utilized for data augmentation. Pedestrian Synthesis GAN (PS-GAN) \cite{PSGAN} generates new pedestrian images to enrich the dataset and stably improve the pedestrian detection model. Similarly, Vobeck\'{y} \textit{et al}. \cite{DataAugmentationForAD} propose a GAN-based framework to augment the pedestrian dataset in autonomous driving. Moreover, they also take human poses as another input to synthesize persons with required poses.

\subsection{Virtual Fitting}

\begin{figure}[t]
	\centering
	\includegraphics[width=0.75\textwidth]{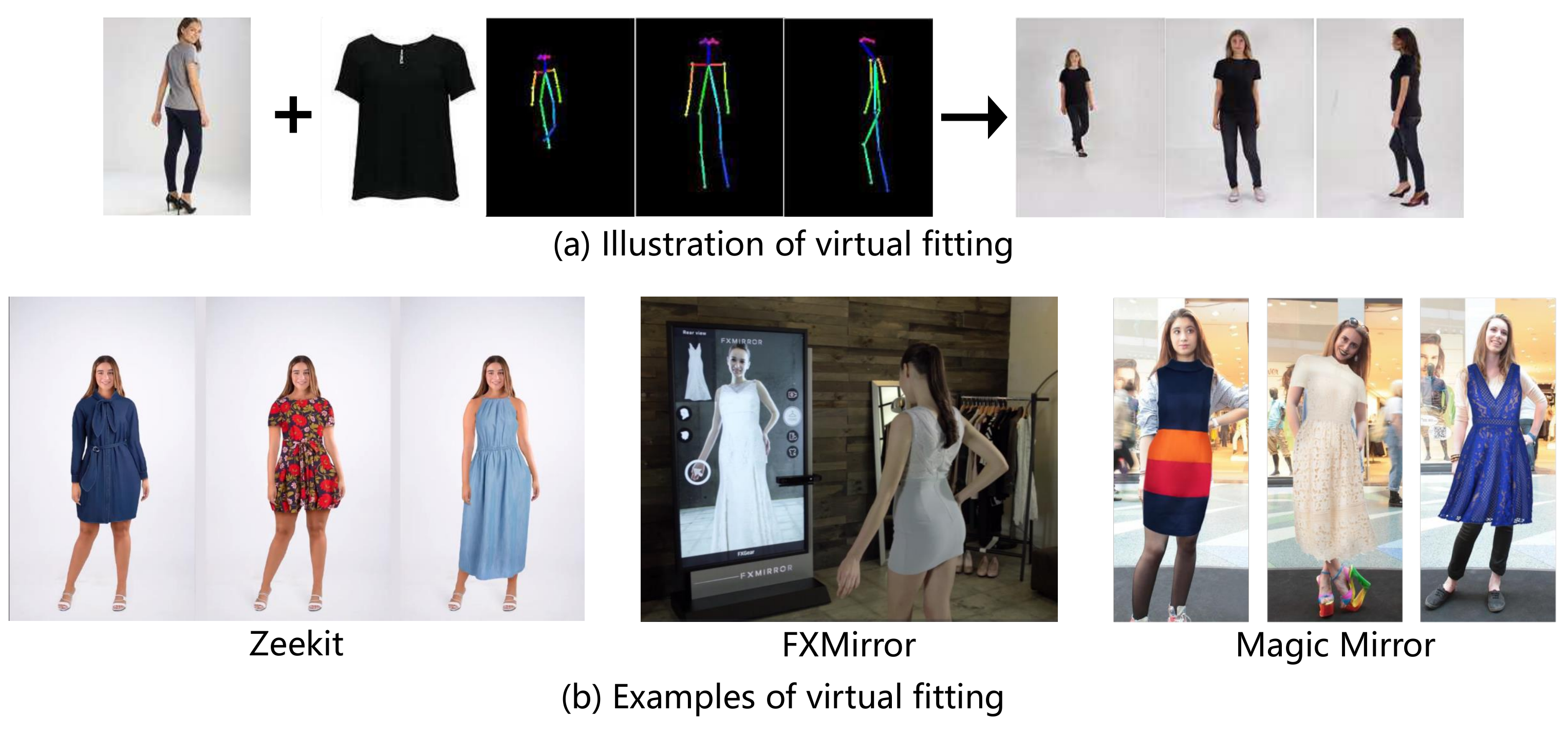}
	\caption{Illustration (a) and example applications (b) of virtual fitting. The target fitting images are controlled by the target cloth and pose.}
	\label{virtual fitting}
\end{figure}

Virtual try-on (Sec.~\ref{HGT}) emphasizes cloth synthesis, while virtual fitting requires both cloth and pose changes in real scenarios, as shown in Fig.\ref{virtual fitting}. For example in online shopping, a big requirement is to generate try-on images based on novel garments and poses. People attempt to combine virtual try-on with pose-guided generation for this purpose.
There have been some virtual fitting apps recently, such as Zeekit\footnote{https://zeekit.me/}, FXMirror\footnote{http://www.fxmirror.net/} and Magic Mirror\footnote{https://www.magicmirror.me/}.

Some approaches are based on top-down methods. FashionOn \cite{FashionOn} transfers the source human parsing into the target pose, and then globally fills the transformed parsing map with clothing textures.
Another stream of works is based on latent feature representation. Yildirim \textit{et al}. \cite{HighResolutionFashion} use latent vectors to represent human pose, garment color and garment parts (e.g., shirt, coat, trousers, skirt, shoes). 
Men \textit{et al}. \cite{ADGAN} meticulously separate a person image into several parts (e.g., pose, head, upper clothes, pants, arms, legs) and then encode them respectively.

Other applications are based on cloth-alignment \cite{FitMe, PGVTON, MGVTON, DownToTheLastDetail}. 
By extending CP-VTON, MG-VTON \cite{MGVTON} deforms the parsing map by the target pose, and takes the target pose as input to generate the result in the target pose.
Wang \textit{et al}. \cite{DownToTheLastDetail} replace standard ResNet Block with tree dilated fusion blocks (tree-blocks) to enrich texture details. Zheng \textit{et al}. \cite{PGVTON} design an attentive bidirectional GAN to highlight clothing textures.
Besides, Dong \textit{et al}. \cite{FWGAN} propose a video-based fitting model (FW-GAN: Flow-navigated Warping GAN) to generate videos with new garments.

\subsection{Digital Human}
\label{DigitalHuman}

\begin{figure}[t]
	\centering
	\includegraphics[width=0.75\textwidth]{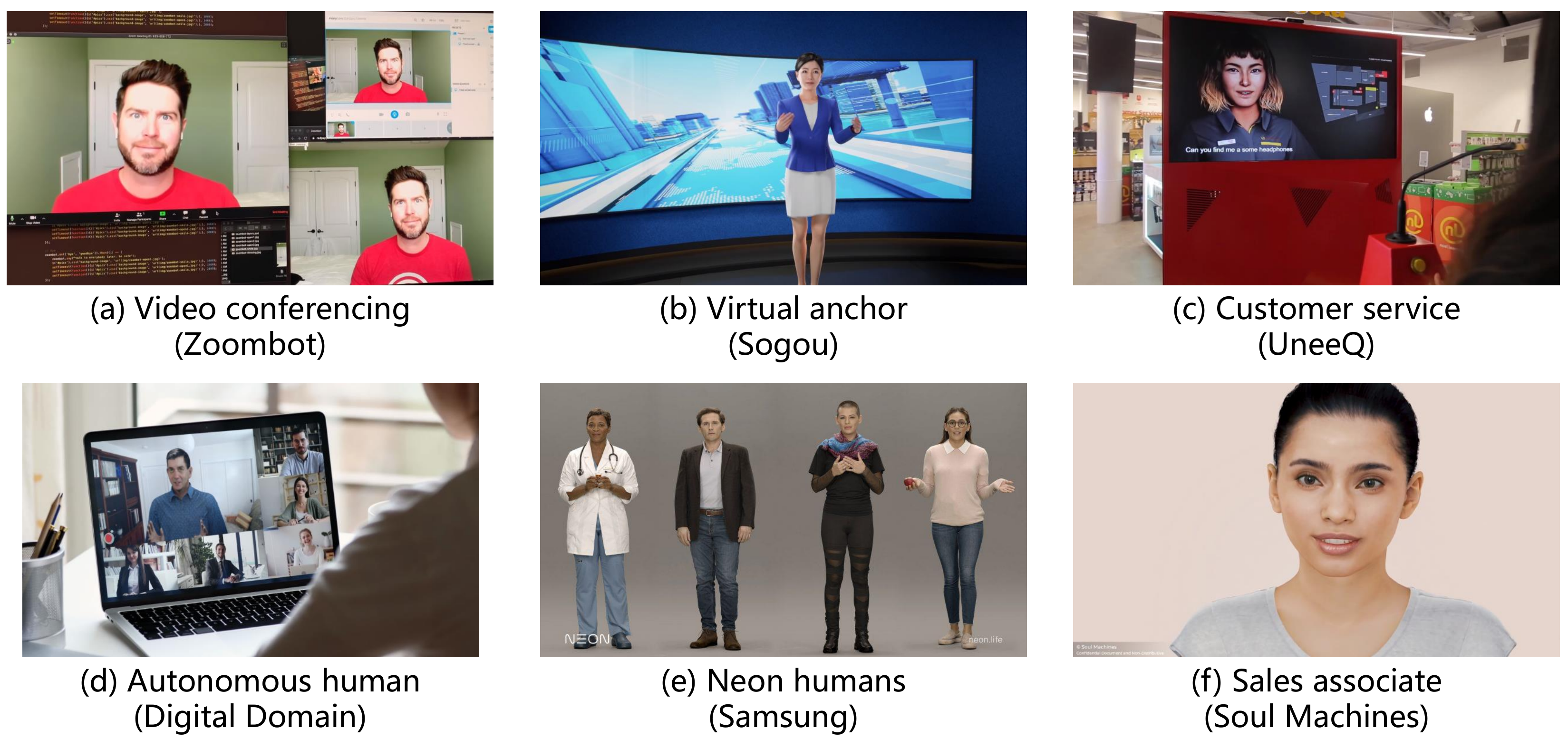}
	\caption{Applications of digital human in different scenarios. The company names are noted in parentheses.}
	\label{digital human}
\end{figure}

Recently, digital human has attracted lots of public attention, which aims to construct a virtual human with realistic appearance and behaviors. Different from robots that have physical body, digital human only exists in the digital realm. At present, digital human has been applied to many scenarios, such as Zoombot video conferencing AI\footnote{https://redpepper.land/blog/zoombot/}, Sogou AI virtual anchor\footnote{https://ai.sogou.com/solution/ai\_host/} and customer service\footnote{https://digitalhumans.com/}, which are shown in Fig.~\ref{digital human}.

A digital human is expected to interact with people with natural language, facial expressions and body gestures, just like real human beings. While a holistic approach is preferred, a feasible way is to combine talking-head and pose-guided generation. Given a speech signal, talking-head generation can synthesize natural lip motions, facial expressions and head movements. Meanwhile, the body can also be synthesized according to the signal to deliver speech-relevant gestures or spontaneous movements \cite{S2GBiLSTM, FastAndSlowS2G, MultiObjS2G, IndividualS2G, End2EndVPF}. The models are trained on large-scale datasets to capture generic speech-to-status (poses, expressions, etc.) mapping and personalized styles.

\section{Future Directions}
\label{Future}

In this paper, we have reviewed deep person generation from three components: face, pose and garment. Thanks to the surprising evolution of deep learning, we have witnessed a rapid development of person generation, from generating low-resolution and rough images to producing high-resolution, detailed and realistic images. However, person generation is still far from mature in generating visually plausible person images/videos on demand. Here we list several future directions worthy of further investigation.

\begin{itemize}

    \item \textbf{Convergence of computer graphics and computer vision.} Computer Graphics (CG) has a mature procedure for creating virtual characters in movies and games. Meanwhile, Computer Vision (CV) has supported the synthesis of photo-realistic human face or body images. Both of them have their unique advantages. Specifically, CG is good at motion control and appearance editing (with explicit mesh, 3DMM, SMPL), and CV produces more realistic appearances (with GAN, Diffusion Model). Recently, there have been some early attempts to combine the advantages of CG and CV, e.g., Neural rendering \cite{SOTAonNR}, NeRF \cite{NeRF} based talking-head generation \cite{ADNeRF}, pose-guided generation based on 3D human mesh \cite{ReenactmentHAV}. For deep person generation, 3D face reconstruction has shown great potential in improving the accuracy of expressions and head motions. 
    
    \item \textbf{Trustworthy contents.} The growing maturity of person generation poses increasing threats to society. Abusing fake person images might cause serious ethical and legal problems, especially in videos of celebrities or politicians. For example, Deepfakes (e.g, head swap, face reenactment) can now produce realistic forged images or videos of celebrities or politicians. 
    Person-related forensics, on the contrary, aims at detecting forged images or videos, which has attracted increasing attention. However, existing works are mostly performance-driven, and thus ignore the model explainability and efficiency. Moreover, most methods are only tuned on a fixed dataset, and the generalization ability is also limited for practical usage. Robust and trustworthy-oriented forgery detection, especially on person-related images or videos, plays an essential role in both accelerating the technical evolution and preventing fake material from being abused.
    
    \item \textbf{Emerging tasks.} Several derived tasks start to emerge with promising prospects. 
    
    \textit{Conversational head generation.} So far, most people focus on how to make digital human speak according to audio and other conditions. Recently, to make more vivid and natural interaction between digital human and real people, Zhou \textit{et al.} \cite{RLHG} propose responsive listening head generation, which allows digital human to respond as a listener, such as nodding and responsive expressions. Listener-centric or even conversational-centric (both talking and listening) generation emerge and has considerable potential for two-way engagement between a virtual agent and human.
    
    \textit{Person-in-context synthesis.} It aims to generate multiple person instances in complex contexts defined by bounding boxes (layout) \cite{PersonInContextSynthesis}. 
    Simply borrowing single-person models for multi-person cases might be sub-optimal. As a more challenging problem, person-in-context synthesis starts to draw attention recently.
    
    \textit{Text-guided person generation.} Text descriptions provide a natural way to interact with humans during generation. Person-related appearance manipulation \cite{TGPIS} is particularly helpful for interactive generation based on human request.
    
\end{itemize}

\begin{acks}
This work was supported in part by the National Natural Science Foundation of China (Grant Nos. 62276017, U1636211, 61672081), the 2022 Tencent Big Travel Rhino-Bird Special Research Program, and the Fund of the State Key Laboratory of Software Development Environment (Grant No. SKLSDE-2021ZX-18).
\end{acks}

\bibliographystyle{ACM-Reference-Format}
\bibliography{sample-base}

\end{document}